 \documentclass[10pt, conference]{IEEEtran}
\IEEEoverridecommandlockouts

\usepackage{cite}
\ifCLASSINFOpdf
  \usepackage[pdftex]{graphicx}
\else
  \usepackage[dvips]{graphicx}
\fi
\usepackage[cmex10]{amsmath}
\usepackage{amssymb}

\usepackage{algorithmic,algorithm}
\usepackage{array}
\usepackage{mdwmath}
\usepackage{mdwtab}
\usepackage[font=footnotesize]{subfig}
\begin{document}
\title{Clustering on Multiple Incomplete Datasets \\via Collective Kernel Learning}

%\author{\IEEEauthorblockN{Michael Shell}
%\IEEEauthorblockA{School of Electrical and\\Computer Engineering\\
%Georgia Institute of Technology\\
%Atlanta, Georgia 30332--0250\\
%Email: http://www.michaelshell.org/contact.html}
%\and
%\IEEEauthorblockN{Homer Simpson}
%\IEEEauthorblockA{Twentieth Century Fox\\
%\and
%\IEEEauthorblockN{James Kirk\\ and Montgomery Scott}
%\IEEEauthorblockA{Starfleet Academy\\
%San Francisco, California 96678-2391\\
\author{\IEEEauthorblockN{Weixiang Shao}
\IEEEauthorblockA{Department of Computer Science\\
University of Illinois at Chicago\\
Chicago, Illinois 60607-7053\\
Email: wshao4@uic.edu}
\and
\IEEEauthorblockN{Xiaoxiao Shi}
\IEEEauthorblockA{Department of Computer Science\\
University of Illinois at Chicago\\
Chicago, Illinois 60607-7053\\
Email: xshi9@uic.edu}
\and
\IEEEauthorblockN{Philip S. Yu}
\IEEEauthorblockA{Department of Computer Science\\
University of Illinois at Chicago\\
Chicago, Illinois 60607-7053\\
Email: psyu@uic.edu}}


%Telephone: (800) 555--1212, Fax: (888) 555--1212}


\maketitle

\begin{abstract}
%\boldmath
Multiple datasets containing different types of features may be available for a given task. For instance,
users' profiles can be used to group users for recommendation systems. 
In addition, a model can also use users' historical behaviors and credit history to group users. 
Each dataset contains different information and suffices for learning.  
A number of clustering algorithms on multiple datasets were proposed during the past few years.
These algorithms assume that at least one dataset is complete. 
So far as we know, all the previous methods will not be applicable if there is no complete dataset available.
However, in reality, there are many situations where no dataset is complete.
As in building a recommendation system, some new users may not have profile or historical behaviors, 
while some may not have credit history. 
Hence, no available dataset is complete.
In order to solve this problem, we propose an approach called \textbf{Co}llective \textbf{K}ernel \textbf{L}earning 
to infer hidden sample similarity from multiple incomplete datasets. 
The idea is to collectively completes the kernel matrices of incomplete datasets 
by optimizing the alignment of shared instances of the datasets. 
Furthermore, a clustering algorithm is proposed based on the kernel matrix.
%In order to solve this problem of incompleteness, we propose an approach called \textbf{Co}llective \textbf{K}ernel \textbf{L}earning, 
%which collectively completes the kernel matrices of incomplete datasets by optimizing the alignment of shared instances of the datasets. 
%We also propose a clustering algorithm based on \textbf{Co}llective \textbf{K}ernel \textbf{L}earning and Kernel Canonical Correlation Analysis.
The experiments on both synthetic and real datasets demonstrate the effectiveness of the proposed approach. 
The proposed clustering algorithm outperforms the comparison algorithms by as much as two times in normalized mutual information.
\end{abstract}


 \ifCLASSOPTIONpeerreview
 \begin{center} \bfseries EDICS Category: 3-BBND \end{center}
 \fi
\IEEEpeerreviewmaketitle

\section{Introduction}
\begin{figure*}[!t]
\centerline
{
\subfloat[User profile. User3 doesn't complete her profile.]
{
\begin{tabular}[b]{l c c c }
    & Name & Age & Country\\
    \begin{minipage}{0.06\columnwidth}
      \includegraphics[width=\columnwidth]{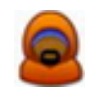}
    \end{minipage}User1 & Bob & 31 & USA\\
    \begin{minipage}{0.06\columnwidth}
      \includegraphics[width=\columnwidth]{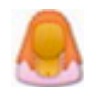}
    \end{minipage}User2 & Angeli & 21 & USA\\
    \begin{minipage}{0.06\columnwidth}
      \includegraphics[width=\columnwidth]{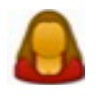}
    \end{minipage}User3 & ? & ? & ?\\
    \begin{minipage}{0.06\columnwidth}
      \includegraphics[width=\columnwidth]{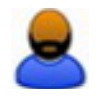}
    \end{minipage}User4 & Tom & 40 & Canada\\
  \end{tabular}
  \label{fig:profile}
}
\hfil
\subfloat[Browsing behaviors. User2 and User4 are new users, thus no browsing behaviors are available.]
{
\begin{tabular}[b]{l c c c }
    & 
    \begin{minipage}{0.16\columnwidth}
      \includegraphics[width=\columnwidth]{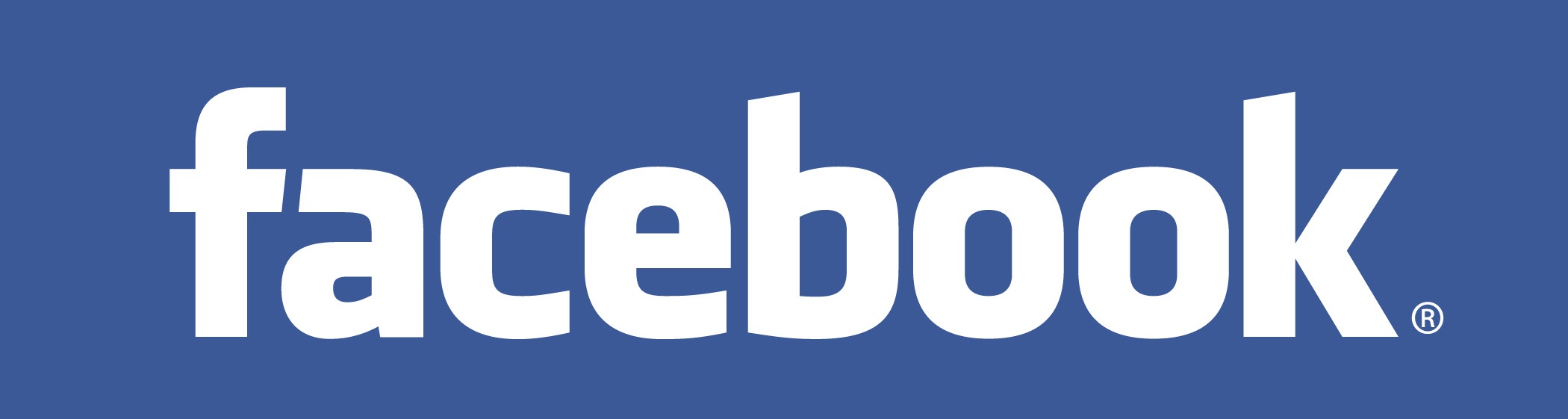}
    \end{minipage} 
     & 
     \begin{minipage}{0.16\columnwidth}
      \includegraphics[width=\columnwidth]{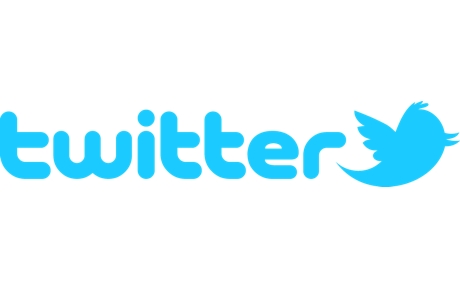}
    \end{minipage} 
     & \begin{minipage}{0.16\columnwidth}
      \includegraphics[width=\columnwidth]{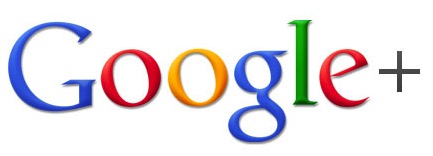}
    \end{minipage} 
    \\
    \begin{minipage}{0.06\columnwidth}
      \includegraphics[width=\columnwidth]{bob.png}
    \end{minipage}User1 & 
    \begin{minipage}{0.06\columnwidth}
      \includegraphics[width=\columnwidth]{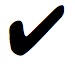}
    \end{minipage}
     & 
     \begin{minipage}{0.06\columnwidth}
      \includegraphics[width=\columnwidth]{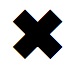}
    \end{minipage}
     & 
     \begin{minipage}{0.06\columnwidth}
      \includegraphics[width=\columnwidth]{cross.png}
    \end{minipage}\\
    \begin{minipage}{0.06\columnwidth}
      \includegraphics[width=\columnwidth]{angeli.png}
    \end{minipage}User2 & 
    ? & ? &?\\
    \begin{minipage}{0.06\columnwidth}
      \includegraphics[width=\columnwidth]{lily.png}
    \end{minipage}User3 & 
    \begin{minipage}{0.06\columnwidth}
      \includegraphics[width=\columnwidth]{check.png}
    \end{minipage}
     & 
     \begin{minipage}{0.06\columnwidth}
      \includegraphics[width=\columnwidth]{cross.png}
    \end{minipage}
     & 
     \begin{minipage}{0.06\columnwidth}
      \includegraphics[width=\columnwidth]{check.png}
    \end{minipage}\\
    \begin{minipage}{0.06\columnwidth}
      \includegraphics[width=\columnwidth]{tom.png}
    \end{minipage}User4 & ? & ? &?\\
    \end{tabular}
  \label{fig:behavior}
}
\hfil
\subfloat[Credit scores. No credit scores for User3 and User4.]
{
\begin{tabular}[b]{c c }
    & Credit Score\\
    \begin{minipage}{0.06\columnwidth}
      \includegraphics[width=\columnwidth]{bob.png}
    \end{minipage}User1 & Fair\\
    \begin{minipage}{0.06\columnwidth}
      \includegraphics[width=\columnwidth]{angeli.png}
    \end{minipage}User2 & Good\\
    \begin{minipage}{0.06\columnwidth}
      \includegraphics[width=\columnwidth]{lily.png}
    \end{minipage}User3 & ?\\
    \begin{minipage}{0.06\columnwidth}
      \includegraphics[width=\columnwidth]{tom.png}
    \end{minipage}User4 & ?\\
  \end{tabular}
  \label{fig:credit}
}
}
\caption{Different datasets for grouping the users in the recommendation systems.}
\label{fig:example}
\end{figure*}

In many real world data mining problems, the same instance may appear in different datasets with different representations.
Different datasets may emphasize different aspects of instances.
An example is grouping the users in an user-oriented recommendation system. 
For this task, related datasets can be (1) user profile database (as shown in Fig.~\ref{fig:profile}),
(2) users' log data (as shown in Fig.~\ref{fig:behavior}), and (3) users' credit score (as shown in Fig.~\ref{fig:credit}).
Learning with such type of data is commonly referred to as multiview learning \cite{Blum:1998:CLU:279943.279962, DBLP:conf/icdm/BickelS04}.
Although there are some previous works on multiple datasets, all of them assume the completeness of the different datasets. 
As far as we know, even the most recently work requires at least one dataset is complete \cite{trivedimultiview}. 
However, in the real world applications, there are many situations in which complete datasets are not available. %%(examples here)
For instance, in Fig.~\ref{fig:profile}, User3 does not complete her profile. However, she has browsing log recorded by the browser. In Fig. \ref{fig:behavior}, checks and crosses indicates whether user visited the website recently. From the figure, we can see that 
User2 and User4 do not have browsing behavior history. This may because that they are new users to the system or 
they refuse to share the historical behaviors with the system.  
In Fig. \ref{fig:credit}, only User1 and User2 have credit scores in the system. In the situation as shown in Fig.~\ref{fig:example}, all the previous method will not be applicable. 
It is very important to find an approach that can work for incomplete datasets.

In order to deal with the incompleteness of the datasets, 
it is a natural way to complete the original datasets first. 
However, it is very hard and time-consuming to directly predict the missing features in each dataset especially if there are large number of missing features. 
Instead, we propose an approach called \textbf{Co}llective \textbf{K}ernel \textbf{L}earning (CoKL). 
This approach iteratively completes the kernel matrix of each dataset using the kernel of other datasets.
Basically, CoKL is based on aligning the similarities between examples across all datasets. 
The completed kernel matrices can be used in any kernel based clustering algorithms.
In this paper, we also propose a clustering algorithm based on CoKL and Kernel Canonical Correlation Analysis (KCCA). 
The proposed clustering algorithm first uses CoKL to complete the kernel matrices. 
Based on the completed kernel matrices, KCCA could find the projections that maximize the correlations between the datasets. 
Then we can perform any
standard clustering algorithms on the projected space.
As compared with previous papers, this paper has several advantages:
\begin{enumerate}
\item The proposed clustering algorithm can be used in situations even when all the datasets are incomplete, in which the other methods are not applicable.
\item Collective kernel learning does not require predicting the missing features in the incomplete datasets using complex method. 
Predicting the missing features may be very time-consuming when there are large number of missing features. 
Instead, we construct the full kernel matrices corresponding to the incomplete datasets iteratively using the shared examples between different datasets.
We only need to give initial values to the missing features in the incomplete datasets to get initial kernel matrices for incomplete datasets.
\end{enumerate}
In order to evaluate the quality of CoKL and the proposed clustering algorithm that uses CoKL and KCCA, 
we conduct several experiments on the UCI seeds data \cite{dataset:seeds} and handwritten Dutch numbers recognition data \cite{dataset:digit}. 
The proposed clustering algorithm outperforms the comparison algorithms by as much as two times in normalized mutual information. 
The experiment on the convergence of CoKL shows that CoKL converges quickly in all the experiment settings (less than 10 iterations). 
Further experiment shows that the number of iterations needed to convergence does not change too much for different missing rates. 

The rest of this paper is organized as follows: 
In the next section, we will describe the formulation of the problem. 
In section \ref{sec:solution}, we will describe the proposed collective kernel learning.
CCA and KCCA are introduced and the clustering algorithm using CoKL and KCCA is described in section \ref{sec:clustering}.
Experiment settings and result analysis are described in section \ref{sec:experiments}. 
The results on different data settings show that the proposed clustering algorithm outperforms the comparison algorithms.
\section{Problem Fomulation}
\label{sec:fomulation}
Before we describe the formulation of the problem, we summarize some notations used in this paper in Table~\ref{tab:notations}.
\newcommand{\tabincell}[2]{\begin{tabular}{@{}#1@{}}#2\end{tabular}}
\begin{table*}[t]
  \centering
  \caption{Notations used in this paper.}
   \begin{tabular}{| l | l |}
   \hline
      \multicolumn{1}{|c|}{Notation} & \multicolumn{1}{c|}{Description}\\
      \hline
      \quad $X$ and $Y$& \quad Incomplete datasets\\
      \hline
      \quad $\mathcal{C} = \{(x_1,x_2),...,(x_c,y_c)\}$ & 
      \quad \tabincell{l}{The set of examples with features present in both $X$ and $Y$. \\ $c$ is the set size.}\\
      \hline
      \quad $\mathcal{M}_1 = \{x_{c+1},...,x_{c+m_1}\}$ &
      \quad \tabincell{l}{The set of examples with features only present in dataset $X$. \\ $m_1$ is the set size.}\\
      \hline
      \quad $\mathcal{M}_2 = \{y_{c+m_1+1},...,y_{c+m_1+m_2}\}$ &
      \quad \tabincell{l}{The set of examples with features only present in dataset $Y$. \\ $m_2$ is the set size.}\\
      \hline
      \quad $K_x$ & \quad A full kernel matrix of data set X with dimension $(c+m_1+m_2)\times(c+m_1+m_2)$. \\
      \hline
      \quad $K_y$ & \quad A full kernel matrix of data set Y with dimension $(c+m_1+m_2)\times(c+m_1+m_2)$.\\
      \hline
      \quad $k(x_i,x_j)$ & \quad The kernel similarity between two examples $x_i$ and $x_i$\\
      \hline
      \quad $\mathcal{L}_x = D_x - K_x$ & \quad \tabincell{l}{The graph Laplacian of Kernel $K_x$,
      where $D_x$ is the diagonal matrix \\consisting of the row sums of $K_x$.} \\
      \hline
      \quad $\mathcal{L}_y = D_y - K_y$ & \quad \tabincell{l}{The graph Laplacian of Kernel $K_y$,
      where $D_y$ is the diagonal matrix \\consisting of the row sums of $K_y$.} \\
      \hline
      \quad $w_x = X\alpha$ & \quad \tabincell{l}{The projection directions for $X$ in CCA problem.\\ Here, $\alpha$ is a vector of size N}\\
      \hline
      \quad $w_y = Y\beta$ & \quad \tabincell{l}{The projection directions for $Y$ in CCA problem.\\ Here, $\beta$ is a vector of size N}\\
      \hline
      \quad $\phi$ & \quad The mapping function that maps a lower dimension data into higher dimension space.\\
      \hline
  \end{tabular}
  \label{tab:notations}
\end{table*}

Given two related datasets $X$ and $Y$, we assume both of these two related datasets are incomplete. 
The features for dataset $X$ are available for only a subset of the total examples, 
and the features for dataset $Y$ are available for another subset of the total examples. 
We also assume these two datasets can cover all the examples, i.e., here are no examples that are missing in both datasets. 
The goal is to derive a clustering solution $\mathcal{S}$ based on both datasets.
Since both datasets are incomplete, 
we denote $\mathcal{C} = \{(x_1, y_1),..., (x_c, y_c)\}$ as the set of examples with features present in both $X$ and $Y$,
$\mathcal{M}_1 = \{x_{c+1},...,x_{c+m_1}\}$ as the set of examples with features only present in dataset $X$, 
and 
$\mathcal{M}_2 = \{x_{c+m_1+1},...,x_{c+m_1+m_2}\}$ as the set of examples with feature only present in dataset $Y$.
So we can rewrite examples in these two datasets as:
\begin{equation}
\notag
X = \begin{pmatrix}
  X_c \\
  X_{m_1}\\
  X_{m_2} = ?
 \end{pmatrix} 
 \qquad
 Y = \begin{pmatrix}
  Y_c \\
  Y_{m_1} = ?\\
  Y_{m_2}
 \end{pmatrix}.
 \end{equation}
 Then we can denote $K_x$, a $(c+m_1+m_2)\times (c+m_1+m_2)$ matrix, 
 as kernel matrix defined over all the examples using features from dataset $X$. 
 The corresponding graph Laplacian \cite{2136896, Spielman_algorithms_graph} is defined as $\mathcal{L}_x = D_x - K_x$,
 where $D_x$ is the diagonal matrix consisting of the row sums of $K_x$ along it's diagonals. 
 Likewise, for dataset $\mathcal{Y}$, we denote the kernel matrix by $K_y$,
 and the corresponding graph Laplacian by $\mathcal{L}_y = D_y - K_y$. 
 However, since features for both $\mathcal{X}$ and $\mathcal{Y}$ are only available for a subset of the total examples, 
 only 4 subblock of the full kernel matrix $K_x$ ($K_y$) with size $c \times c$, $c \times m_1$, $m_1 \times c$, $m_1 \times m_1$
  ($c \times c$, $c \times m_2$, $m_2 \times c$, $m_2 \times m_2$) will be available 
 (see Equation~\ref{equation:incomplete_kernels}). 
 \begin{equation}
 \label{equation:incomplete_kernels}
 \begin{split}
 K_x = & \begin{pmatrix}
 K_x^{cc} & K_x^{cm_1} & K_x^{cm_2} = ? \\
 (K_x^{cm_1})^T & K_x^{m_1m_1} & K_x^{m_1m_2} = ?\\
 (K_x^{cm_2})^T=? & (K_x^{m_1m_2})^T=? & K_x^{m_2m_2} = ?
 \end{pmatrix}\\
 K_y = & \begin{pmatrix}
 K_y^{cc} & K_y^{cm_1}=? & K_y^{cm_2} \\
 (K_y^{cm_1})^T = ? & K_y^{m_1m_1} = ? & K_y^{m_1m_2} = ?\\
 (K_y^{cm_2})^T & (K_y^{m_1m_2})^T=? & K_y^{m_2m_2}
 \end{pmatrix}
\end{split}
\end{equation}
In order to apply any kernel approach for clustering, one must first build the full kernel matrix $K_x$ and $K_y$. 
To achieve this goal, we borrow the idea from Laplacian regularization \cite{DBLP:conf/colt/SmolaK03, NIPS2006_662, NIPS2009_0792}. 
In other words, 
we first generate the graph Laplacian $\mathcal{L}_x$ for the kernel matrix $K_x$. 
Then $tr(\mathcal{L}_xK_y)$ reflects the ``inconsistence'' of the kernel matrix $K_y$ 
when we ``explain'' it with the graph Laplacian $\mathcal{L}_x$ from $K_x$. 
In this paper, $tr$ denotes the matrix trace.
Under the assumption that $K_x$ and $K_y$ should contain consensus information, 
we should minimize the ``inconsistence" $tr(\mathcal{L}_xK_y)$, 
and similarly for  $tr(\mathcal{L}_yK_x)$. 
More formally, the objective can be written as follows:
%In order to apply any kernel approach for clustering, one must first build the full kernel matrix $K_x$ and $K_y$.
%Using the ideas from Laplacian regularization, this can be approximatively achieved by 
%solving the following two optimization problems:
\begin{align}
&\min_{K_y \succeq 0} tr(\mathcal{L}_x K_y) \label{equation:opt1} \\
&\min_{K_x \succeq 0} tr(\mathcal{L}_y K_x) \label{equation:opt2}\\
s.t.\qquad&K_y(i_1,j_1) = k(y_{i_1},y_{j_1}), \notag\\
&\text{where }i_1\text{ and }j_1 \text{ are instances in dataset } Y.\notag\\ 
&K_x(i_2,j_2) = k(x_{i_2},x_{j_2}), \notag\\
&\text{where }i_2\text{ and }j_2 \text{ are instances in dataset } X .\notag
\end{align}

Here, $k(y_{i_1},y_{j_1})$ is the kernel similarity between two examples $y_{i_1}$ and $y_{i_2}$ in dataset $Y$, and
$k(x_{i_2},x_{j_2})$ is the kernel similarity between two examples $x_{i_1}$ and $x_{i_2}$ in dataset $X$.
The objective functions above optimize the alignment between $K_x$ and $K_y$, 
given the known part of $K_x$ and $K_y$. 
Now we only need to solve the above optimization problems.

\section{Collective Kernel Learning}
\label{sec:solution}
To construct the full kernel matrices for incomplete datasets, we need to solve the optimization problems in Equations~\ref{equation:opt1} and \ref{equation:opt2}.
However, optimizing Equation~\ref{equation:opt1} requires the completeness of $K_x$, 
and optimizing Equation~\ref{equation:opt2} requires the completeness of $K_y$.
Since both datasets are incomplete, none of $K_x$ and $K_y$ is complete.
We could not just solve these optimization problems directly.
However, we could use collective kernel learning to approximately solve the problem.
We can first fix one of the kernel matrix, 
say fix $K_x$ by giving the missing features in dataset $\mathcal{X}$ initial guesses to construct the initial full kernel matrix $K_x$.
Then we can optimize $K_y$ by solving one of the two optimization problems 
$\min_{K_y \succeq 0} tr(\mathcal{L}_x K_y)$.
Using the completed kernel matrix $K_y$, we can optimize $K_x$ by solving $\min_{K_x \succeq 0} tr(\mathcal{L}_y K_x)$.
This optimization process can continue until it converges.

Without loosing generality, we first fix $K_x$ (filling the missing features in $X$ with average values for continuous features and majority values for discrete  features) 
and use $K_x$ to solve the optimization problem in Equation \ref{equation:opt1}. 
Since both the kernel matrices $K_x$ and $K_y$ should satisfy the positive semi-definite constraint, 
we can express $K_y$ as $AA^T$ (or $K_x$ as $BB^T$), where $A$ (or $B$) is a matrix of real numbers.
Let us write $A$ as $A = \begin{pmatrix} A_c \\ A_{m_1}\\ A_{m_2}\end{pmatrix}$,
and $\mathcal{L}_x$ as:
$$
\mathcal{L}_x = 
\begin{pmatrix} 
\mathcal{L}_x^{cc} & \mathcal{L}_x^{cm_1} & \mathcal{L}_x^{cm_2}\\
(\mathcal{L}_x^{cm_1})^T & \mathcal{L}_x^{m_1m_1} & \mathcal{L}_x^{m_1m_2}\\
(\mathcal{L}_x^{cm_2})^T &(\mathcal{L}_x^{m_1m_2})^T & \mathcal{L}_x^{m_2m_2}
\end{pmatrix}.
$$
Using these and the property of trace, we can rewrite Equation \ref{equation:opt1} as follows:
\begin{equation}
\begin{split}
&\min_Atr\left( \mathcal{L}_xAA^T\right) = \min_Atr\left(A^T\mathcal{L}_xA\right) \\
&= \min_{A_c, A_{m_1}, A_{m_2}}tr \left( \begin{pmatrix}A_c\\A_{m_1}\\A_{m_2}\end{pmatrix}^T \times \right. \\
&\left.\begin{pmatrix} 
\mathcal{L}_x^{cc} & \mathcal{L}_x^{cm_1} & \mathcal{L}_x^{cm_2}\\
(\mathcal{L}_x^{cm_1})^T & \mathcal{L}_x^{m_1m_1} & \mathcal{L}_x^{m_1m_2}\\
(\mathcal{L}_x^{cm_2})^T &(\mathcal{L}_x^{m_1m_2})^T & \mathcal{L}_x^{m_2m_2}
\end{pmatrix} \times
\begin{pmatrix}A_c\\A_{m_1}\\A_{m_2}\end{pmatrix} \right).
\end{split}
\end{equation}

Expanding the above, and using the fact that $A_c$ and $A_{m_2}$ are constant 
(since $A_cA_c^T = K_y^{cc}$ and $A_{m_2}A_{m_2}^T = K_y^{m_2m_2}$ are constant), we get:
\begin{equation}
\begin{split}
&\min_{A_{m_1}} tr(A_c^T\mathcal{L}_x^{cc}A_c + A_{m_1}^T(\mathcal{L}_x^{cm_1})^TA_c+A_{m_2}^T(\mathcal{L}_x^{cm_2})^TA_c+\\
& A_c^T\mathcal{L}_x^{cm_1}A_{m_1} + A_{m_1}^T\mathcal{L}_x^{m_1m_1}A_{m_1} + A_{m_2}^T(\mathcal{L}_x^{m_1m_2})^TA_{m_1} + \\
& A_c^T\mathcal{L}_x^{cm_2}A_{m_2} + A_{m_1}^T\mathcal{L}_x^{m_1m_2}A_{m_2} + A_{m_1}^T\mathcal{L}_x^{m_2m_2}A_{m_2}).
\end{split}
\notag
\end{equation}
Since for any matrix $X$, $tr(X) = tr(X^T)$, we can simplify the above as:
\begin{equation}
\begin{split}
&\min_{A_{m_1}} tr( A_cA_c^T\mathcal{L}_x^{cc}) + 2tr(A_{m_1}A_cT\mathcal{L}_x^{cm_1}) + \\
&2 tr(A_{m_2}A_c^T\mathcal{L}_x^{cm_2})+ 2tr(A_{m_2}A_{m_1}^T\mathcal{L}_x^{m_1m_2}) + \\
&tr(A_{m_1}A_{m_1}^T\mathcal{L}_x^{m_1m_1}) + tr(A_{m_2}A_{m_2}^T\mathcal{L}_x^{m_2m_2}).
\end{split}
\notag
\end{equation}  
Using the fact that $A_cA_c^T = K_y^{cc}$ and $A_{m_2}A_{m_2}^T = K_y^{m_2m_2}$ are constant, 
we can further simplify the above by removing $tr( A_cA_c^T\mathcal{L}_x^{cc})$ and $tr(A_{m_2}A_{m_2}^T\mathcal{L}_x^{m_2m_2})$:
\begin{equation}
\begin{split}
\min_{A_{m_1}} & 2tr(A_{m_1}A_c^T\mathcal{L}_x^{cm_1}) + 2tr(A_{m_2}A_c^T\mathcal{L}_x^{cm_2})+\\
&2tr(A_{m_2}A_{m_1}^T\mathcal{L}_x^{m_1m_2}) + tr(A_{m_1}A_{m_1}^T\mathcal{L}_x^{m_1m_1}).
\end{split}
\notag
\end{equation}  
Taking derivative w.r.t. $A_{m_1}$ and setting it to zero, we get:
\begin{equation}
\begin{split}
2(\mathcal{L}_x^{cm_1})^TA_c + 2\mathcal{L}_x^{m_1m_2}A_{m_2} + 2 \mathcal{L}_x^{m_1m_1}A_{m_1} = 0\\
\end{split}.
\notag
\end{equation} 
Solve the equation, we get:
\begin{equation}
A_{m_1} = -(\mathcal{L}_x^{m_1m_1})^{-1}\left((\mathcal{L}_x^{cm_1})^TA_c - \mathcal{L}_x^{m_1m_2}A_{m_2}\right).
\end{equation}
Thus
\begin{equation}
\notag
\begin{split}
A &= \begin{pmatrix} 
A_c\\
A_{m_1}\\
A_{m_2}
\end{pmatrix} \\
&= \begin{pmatrix} 
A_c\\
-(\mathcal{L}_x^{m_1m_1})^{-1}\left((\mathcal{L}_x^{cm_1})^TA_c - \mathcal{L}_x^{m_1m_2}A_{m_2}\right)\\
A_{m_2}
\end{pmatrix}.
\end{split}
\end{equation}
Then using $K_y = AA^T$, $A_cA_c^T = K_y^{cc}$, $A_{c}A_{m_2}^T = K_y^{cm_2}$, $A_{m_2}A_{m_2} = K_y^{m_2m_2}$, we get:
\begin{equation}
\label{equation:solution}
\begin{split}
K_y = & \begin{pmatrix}
 K_y^{cc} & K_y^{cm_1} & K_y^{cm_2} \\
 (K_y^{cm_1})^T & K_y^{m_1m_1}  & K_y^{m_1m_2}\\
 (K_y^{cm_2})^T & (K_y^{m_1m_2})^T& K_y^{m_2m_2}
 \end{pmatrix}\\
\end{split},
\end{equation}
where 
\begin{eqnarray}
\notag
\begin{split}
K_y^{cm_1} &= -(K_y^{cc}\mathcal{L}_x^{cm_1} + \mathcal{L}_x^{cm_2}(\mathcal{L}_x^{m_1m_2})^T)((\mathcal{L}_x^{m_1m_1})^{-1})^T\\
K_y^{m_1m_1} &=(\mathcal{L}_x^{m_1m_1})^{-1}( (\mathcal{L}_x^{cm_1})^TK_y^{cc} \\
&+ \mathcal{L}_x^{m_1m_2}(K_y^{cm_2})^T\mathcal{L}_x^{cm_1} + (\mathcal{L}_x^{cm_1})^TK_y^{cm_2}(\mathcal{L}_x^{m_1m_2})^T \\
&+ \mathcal{L}_x^{m_1m_2}K_y^{m_2m_2}(\mathcal{L}_x^{m_1m_2})^T)((\mathcal{L}_x^{m_1m_1})^{-1})^T\\
K_y^{m_1m_2} &=  -(\mathcal{L}_x^{m_1m_1})^{-1}((\mathcal{L}_x^{cm_1})^TK_y^{cm_2} + \mathcal{L}_x^{m_1m_2}K_y^{m_2m_2}).
\end{split}
\end{eqnarray}

Similarly, fix $K_y$, solving the optimization in Equation \ref{equation:opt2}, we get:
\begin{equation}
\notag
\begin{split}
K_x = & \begin{pmatrix}
 K_x^{cc} & K_x^{cm_1} & K_x^{cm_2} \\
 (K_x^{cm_1})^T & K_x^{m_1m_1}  & K_x^{m_1m_2}\\
 (K_x^{cm_2})^T & (K_x^{m_1m_2})^T& K_x^{m_2m_2}
 \end{pmatrix}\\
\end{split},
\end{equation}
where
\begin{eqnarray}
\notag
\begin{split}
K_x^{cm_2} &= -(K_x^{cc}\mathcal{L}_y^{cm_2} + \mathcal{L}_y^{cm_1}(\mathcal{L}_y^{m_2m_1})^T)((\mathcal{L}_y^{m_2m_2})^{-1})^T\\
K_x^{m_2m_2} &=(\mathcal{L}_y^{m_2m_2})^{-1}( (\mathcal{L}_y^{cm_2})^TK_x^{cc} \\
&+ \mathcal{L}_y^{m_2m_1}(K_x^{cm_1})^T\mathcal{L}_y^{cm_2} + (\mathcal{L}_y^{cm_2})^TK_x^{cm_1}(\mathcal{L}_y^{m_2m_1})^T \\
&+ \mathcal{L}_y^{m_2m_1}K_x^{m_1m_1}(\mathcal{L}_y^{m_2m_1})^T)((\mathcal{L}_y^{m_2m_2})^{-1})^T\\
K_x^{m_2m_1} &=  -(\mathcal{L}_y^{m_2m_2})^{-1}((\mathcal{L}_y^{cm_2})^TK_x^{cm_1} + \mathcal{L}_y^{m_2m_1}K_x^{m_1m_1}).
\end{split}
\end{eqnarray}
So we can iteratively solve the optimization problems in Equations \ref{equation:opt1} and \ref{equation:opt2} until it gets convergence.
The whole algorithm is shown in Algorithm \ref{algorithm:1}.

Although Algorithm~\ref{algorithm:1} is for two incomplete datasets, it is important to note that 
the generalization can be easily done.  
By  completing the kernel matrices in a cyclic iteration, 
Algorithm~\ref{algorithm:1} can be easily generalized to more than two incomplete datasets . 
Assume we have $k$ incomplete datasets $X_1,...,X_k$.
We first complete the kernel matrix $K_2$ using the initial kernel matrix $K_1$ by Equation~\ref{equation:solution}.
We can continue completing kernel matrix $K_{i+1}$ using kernel matrix $K_i$ by Equation~\ref{equation:solution},
until we complete $K_k$. 
After using $K_k$ to complete $K_1$, we can start another iteration cycle from $K_1$ to $K_k$, until it converges.

\begin{algorithm}
\caption{ Collective Kernel Learning (CoKL)}
\label{algorithm:1}
\begin{algorithmic}[1]
\algsetup{indent=2em}
\renewcommand{\algorithmicrequire}{\textbf{Input:}}
\renewcommand{\algorithmicensure}{\textbf{Output:}}
\REQUIRE Incomplete Datasets $X$ and $Y$
\ENSURE The full kernel matrices $K_x$ and $K_y$\\
\item[]
\STATE Give initial values to the missing features in the two dataset. 
\STATE Calculate the kernel matrices $K_x$ and $K_y$.
\STATE $\mathcal{L}_x \leftarrow  D_x - K_x$ \\
\STATE $\mathcal{L}_y \leftarrow  D_Y - K_y$ \\
\REPEAT
\STATE {
\begin{equation}
\notag
\begin{split}
A & \leftarrow \begin{pmatrix} 
A_c\\
-(\mathcal{L}_x^{m_1m_1})^{-1}\left((\mathcal{L}_x^{cm_1})^TA_c - \mathcal{L}_x^{m_1m_2}A_{m_2}\right)\\
A_{m_2}
\end{pmatrix}
\end{split}
\end{equation}
}
\STATE Calculate the new full kernel matrix $K_y'$ using A.\\
\STATE \begin{equation}
\notag
\begin{split}
B & \leftarrow \begin{pmatrix} 
B_c\\
B_{m_1}\\
-(\mathcal{L}_y^{m_2m_2})^{-1}\left((\mathcal{L}_y^{cm_2})^TB_c - \mathcal{L}_y^{m_2m_1}B_{m_1}\right)
\end{pmatrix}
\end{split}
\end{equation}
\STATE Calculate the new full kernel matrix $K_x'$ using B. \\
\STATE $K_x \leftarrow K_x'$.\\
\STATE $K_y \leftarrow K_y'$.
\UNTIL{Convergence}
\end{algorithmic}
\end{algorithm}

\section{Clustering Algorithm Based on Collective Kernel Learning and KCCA}
\label{sec:clustering}
In this section, we propose a clustering algorithm based on collective kernel learning 
and kernel canonical correlation analysis. 
\subsection{CCA and Kernel CCA}
Canonical Correlation Analysis (CCA) \cite{Johnson:1988:AMS:59551} is a technique for modeling the relationships between two
(or more) sets of variables. CCA computes a low-dimensional shared embedding of both sets of
variables such that the correlations among the variables between the two sets is maximized in the
embedded space.
Given two column vectors $X = (x_1,...,x_n)$ and $Y = (y_1, ..., y_m)$  of random variables with finite second moments, 
canonical correlation analysis seeks vectors $a$ and $b$  such that the random variables $a'X$  and $b'Y$ maximize the correlation $\rho = cor(a'X, b'Y)$. 
CCA has been applied with great success in the past on a variety of learning problems dealing with multi-modal data or multi view data \cite{Chaudhuri:2009:MCV:1553374.1553391}. 
Canonical Correlation Analysis is a linear feature extraction algorithm. 
However, in real world applications, the data usually exhibit nonlinearities, and therefor a linear projection like CCA may not be able to capture the properties of the data. 
To deal with the nonlinearities, kernel method has been successfully used in many applications (e.g. Support Vector Machines and Kernel Principal Component Analysis). 
\cite{lai2000kernel, akaho2006kernel} apply the kernel method to CCA, 
which first maps each $D$ dimensional data point $x$ to a higher dimensional space $\mathcal{F}$ defined by a mapping function $\phi$ whose range is in an inner product space, then applies linear CCA in the feature space $\mathcal{F}$.
More formally, to get the kernel formulation of CCA, 
we can switch to the dual representation by expressing the projection directions as
$w_x = X\alpha$ and $w_y = Y\beta$, where $\alpha$ and $\beta$ are vectors of size N. 
Then the correlation coefficient between $X$ and $Y$ can be written as:
\begin{equation}
\rho = \max_{\alpha, \beta} \frac{\alpha^TX^TXY^TY\beta}{\sqrt{\alpha^TX^TXX^TX\alpha\times\beta^TY^TYY^TY\beta}}.
\end{equation}
Using the fact that $K_x = X^TX$ and $K_y = Y^TY$ are the kernel matrices for $X$ and $Y$, kernel CCA aims to solving the following problem:
\begin{eqnarray}
\rho = \max_{\alpha, \beta} \frac{\alpha^TK_xK_y\beta}{\sqrt{\alpha^TK_x^2\alpha\times\beta^TK_y^2\beta}}\\
\text{s.t. } \alpha^TK_x^2\alpha = 1 \text{ and } \beta^TK_y^2\beta = 1. \notag
\end{eqnarray}
Unlike the linear CCA doing an eigen-decomposition of the covariance matrix, Kernel CCA works by using the kernel metrices $K_x$ and $K_y$.
The eigenvalue problem for kernel CCA is:
\begin{equation}
\begin{pmatrix}
  0 & K_xK_y \\
  K_xK_y & 0\\
 \end{pmatrix} \begin{pmatrix} \alpha\\\beta\end{pmatrix}
 = \lambda \begin{pmatrix}K_x^2 & 0\\ 0& K_y^2\end{pmatrix}\begin{pmatrix}\alpha\\ \beta\end{pmatrix}.
\end{equation}

\subsection{A Clustering Algorithm with Collective Kernel Learning and KCCA}
In this section, we will describe a clustering algorithm based on collective kernel learning and KCCA.
Given two incomplete datasets $X$ and $Y$, 
the goal is to derive a clustering solution $\mathcal{S}$ based on the information contains in both datasets.
The algorithm is shown in Algorithm~\ref{algorithm:2}.
\begin{algorithm}
\caption{Clustering using Collective Kernel Learning and KCCA}
\label{algorithm:2}
\begin{algorithmic}
\renewcommand{\algorithmicrequire}{\textbf{Input:}}
\renewcommand{\algorithmicensure}{\textbf{Output:}}
\REQUIRE Incomplete Datasets $X$ and $Y$.
\ENSURE The clustering solution $\mathcal{S}$.
\item[]
\STATE $[K_x, K_y] = CoKL(X, Y)$.\\
\STATE $[X_p, Y_p] = KCCA(K_x, K_y)$.  \COMMENT{$X_p$ and $Y_p$ are the projected datasets.}
\IF{The feature space is still too large} 
\STATE{$X_p = PCA(X_p)$}. 
\STATE{$Y_p = PCA(Y_p)$}.
\ENDIF
\STATE Apply k-means to the projected datasets $X_p$ and $Y_p$.
\end{algorithmic}
\end{algorithm}

We first apply the collective kernel learning to complete the two full kernel matrices. 
Then we use Kernel CCA to find the projected feature space, in which the correlation of the two datasets is maximized, 
and get the projected two datasets.
In case that the dimension of the projected feature space is still to large for clustering, we apply Principal Component Analysis (PCA) to the projected datasets if needed.
The clustering solution $\mathcal{S}$ can be acquired  using any  standard clustering algorithm, like k-means. 

\section{Experiments and Results}
\label{sec:experiments}
In this section, we analyze the proposed clustering algorithm on two sets of datasets. 
\begin{figure*}[!t]
\centerline{
\subfloat[NMI for seeds dataset on different missing rates.] {
\includegraphics[width=0.47\textwidth]{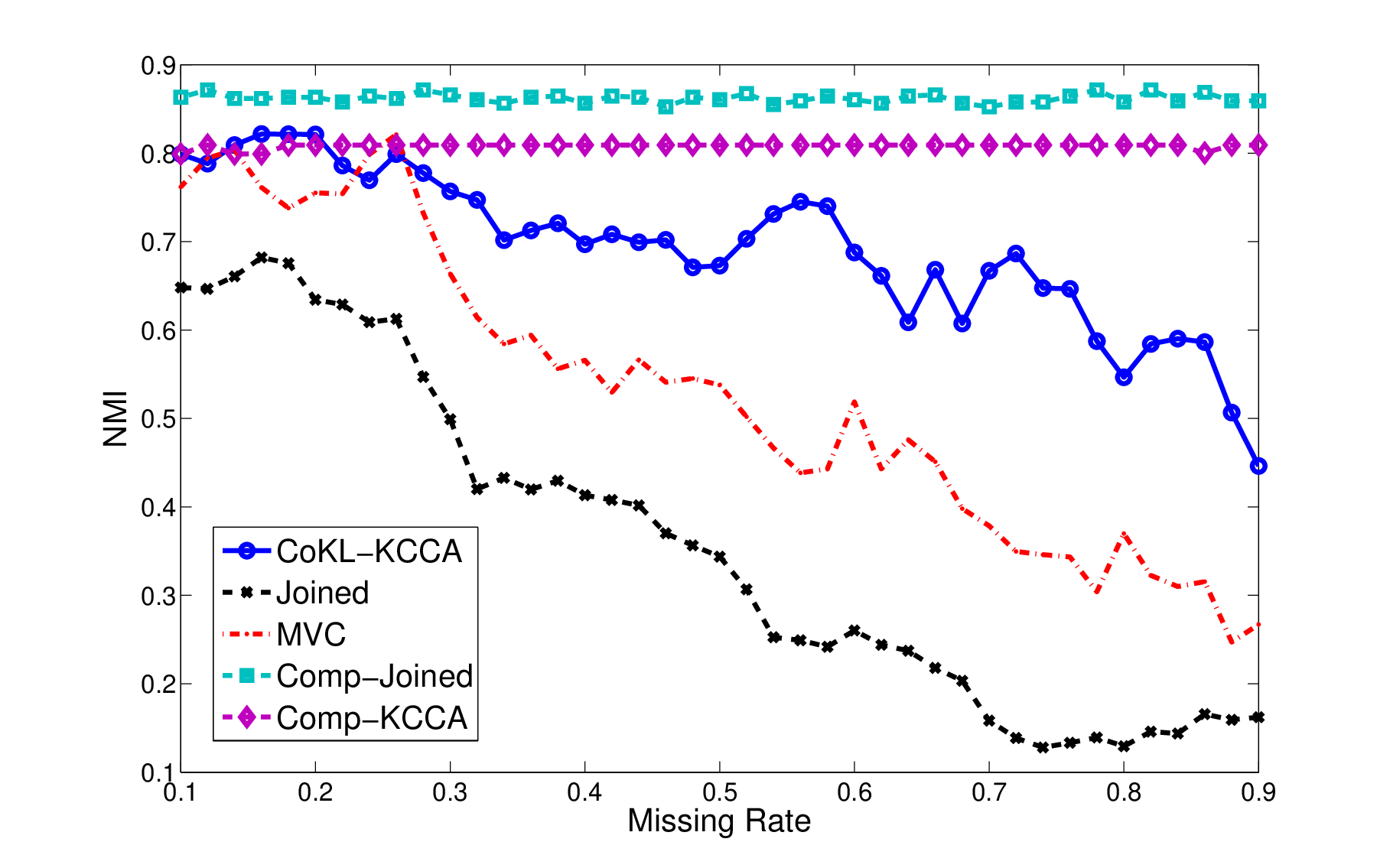}
\label{fig:seeds_1}
}
\subfloat[Average purity for seeds dataset on different missing rates.] {
\includegraphics[width=0.47\textwidth]{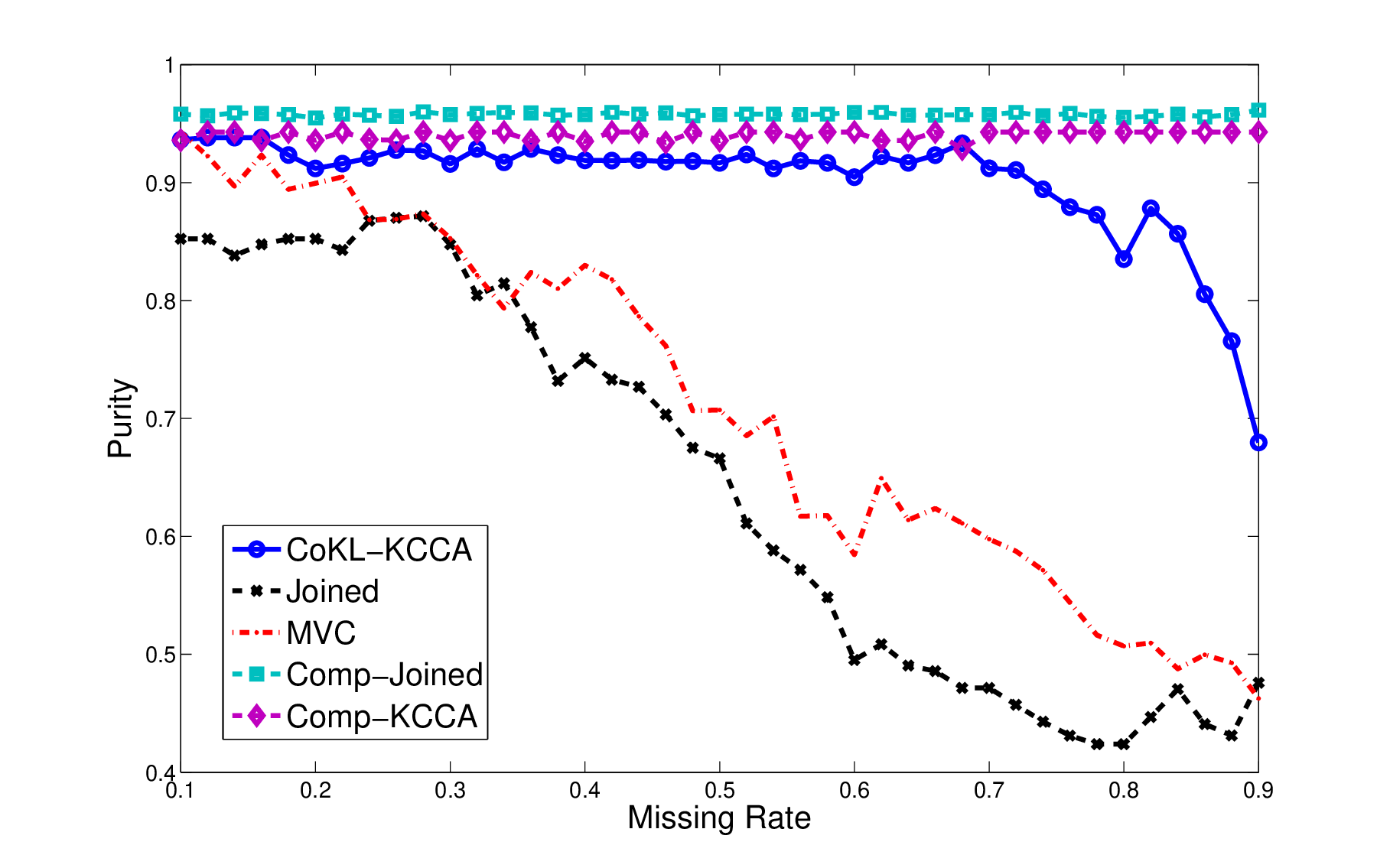}
\label{fig:seeds_2}
}}
\caption{The performance of seeds dataset on different missing rates.}
\label{fig:seeds}
\end{figure*}
\begin{figure*}[!t]
\centerline{
\subfloat[NMI for seeds dataset on different missing rates.] {
\includegraphics[width=0.47\textwidth]{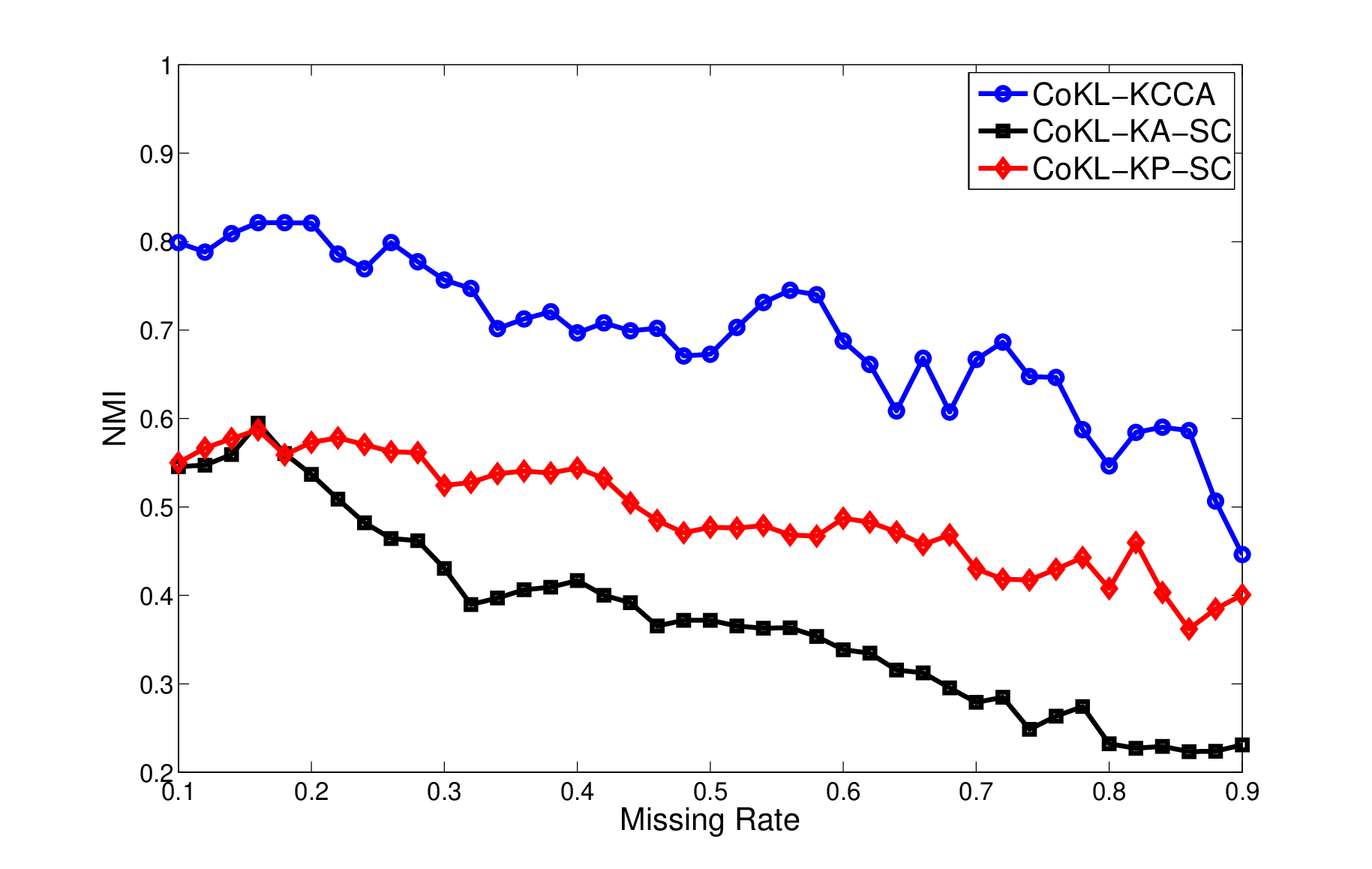}
\label{fig:seeds_3}
}
\subfloat[Average purity for seeds dataset on different missing rates.] {
\includegraphics[width=0.47\textwidth]{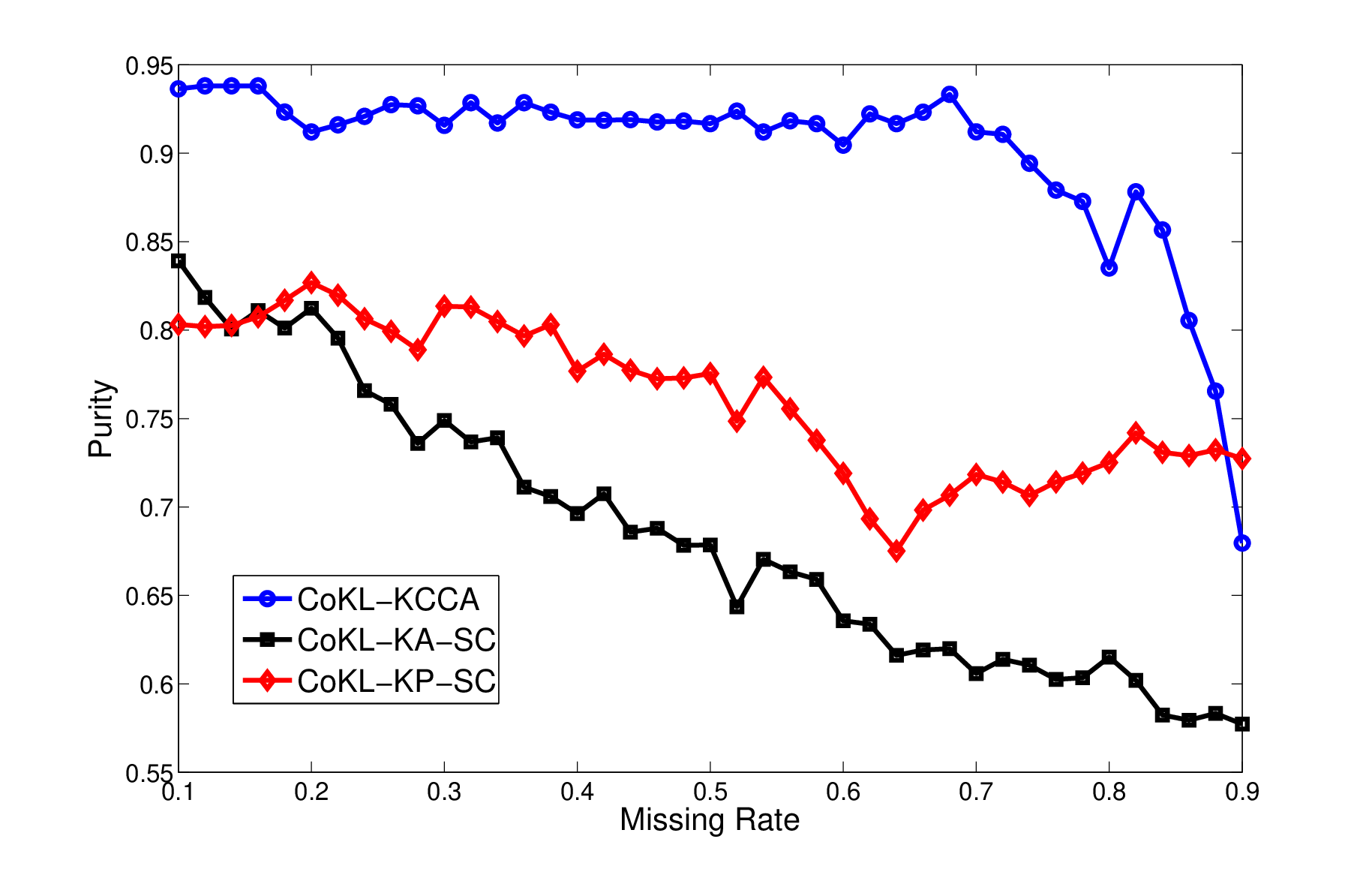}
\label{fig:seeds_4}
}}
\caption{The performance of CoKL with different algorithms on different missing rates.}
\label{fig:seeds2}
\end{figure*}
\subsection{Comparison Approaches}
Since there was no previous method that can be directly used to handle the same problem, 
we compared the proposed algorithm with a straightforward strategy. 
The comparison strategy is to first fill the missing features with average values for continuous  features and majority values for discrete  features, and then concatenate all features together, referred as Concat. 
In other words, given two incomplete datasets $X$ and $Y$, we just fill the missing features and get $X_c$ and $Y_c$. 
The concatenated features can be represented as follows:
\begin{equation}
F_{XY} = [X_c^T, Y_c^T]^T.
\end{equation}
So any traditional clustering algorithm can be applied on the concatenated datasets to obtain a solution. 
Another comparison Approach is the algorithms in \cite{trivedimultiview} referred as MVC. 
This algorithm assumes at least one dataset is complete.
To apply this algorithm to the incomplete datasets, we complete one dataset with average values for continuous features and majority values for discrete features, 
leaving other datasets incomplete. Although the proposed clustering algorithm could work with any standard clustering algorithm, 
in all the experiments, we use k-means as the clustering algorithm for convenience.

To test the effectiveness of KCCA, we compare the proposed algorithm (CoKL+KCCA) with Kernel Addition/Production + Spectral Clustering referred as CoKL-KA-SC and CoKL-KP-SC. 
CoKL-KA-SC is combining different kernels from CoKL by adding them, and then running standard spectral clustering on the corresponding Laplacian.
As suggested in \cite{NIPS2009_0716, NIPS2011_0817}, even this
seemingly simple approach often leads to near optimal results as compared to more sophisticated
approaches for classiÞcation.
CoKL-KP-SC is multiplying the corresponding entries of kernels after CoKL and applying standard spectral clustering on the resultant Laplacian.
To be fair, we also compare the proposed algorithm (CoKL+KCCA) with concatenated standard k-means and KCCA on the complete dataset referred as Comp-Concat and Comp-KCCA. 

\subsection{Evaluation Strategy}
In order to evaluate the quality of the proposed clustering algorithm, we use normalized mutual information (NMI) and the average purity. 
Note that NMI equals to zero when clustering algorithm is random, and it is close to one when the clustering result is good.
Average purity is also close to one when the clustering result is good.
Note that k-means is sensitive to initial seed selection. 
Hence, we run k-means 30 times on each parameter setting, and report the averaged NMI and purity with mean value and standard deviation. 
All the datasets we use in the experiments are complete, but we randomly delete some of the instances in datasets.
It is also important to note that the missing rate for each dataset is equal, i.e., two datasets have the same number of missing instances. 
Since all the original datasets are complete, to generate a missing rate of 60\% on a pair of datasets, 
we randomly select 60\% of the instances and delete them alternately from one of the datasets. 
This will make all the datasets have equal missing rate.
We test the performance of the proposed algorithm for different total missing rates (from 10\% to 90\%).
\subsection{UCI Seeds Datasets}
The first dataset contains 210 instances with 7 features. 
Each instance represents a seed belonging to one of the three different varieties of wheat. 
A soft X-ray technique and GRAINS package are used to construct all seven, 
real-valued attributes. 
The aim is to cluster the seeds.
In order to test the performance of the proposed algorithm,
we randomly split the feature set into two disjoint parts, 
which represent two datasets. 
Then we randomly delete the instances in both of the datasets to make them incomplete.
As mentioned before, we run the proposed algorithm for different total missing rates (from 10\% to 90\%). 
The results average over 30 runs are presented in Fig.~\ref{fig:seeds} and Fig.~\ref{fig:seeds2}.

Fig.~\ref{fig:seeds} compares CoKL+KCCA with Concat, MVC, and two algorithms on complete data (Comp-Concat and Comp-KCCA). 
Fig.~\ref{fig:seeds2} compares different algorithms combined with CoKL on incomplete data (CoKL+KCCA, CoKL-KA-SC and CoKL-KP-SC).
As it can be observed in Fig.~\ref{fig:seeds}, the proposed algorithm, clustering with CoKL and KCCA,
 outperforms the two comparison methods (Concat and MVC) substantially for all the missing rates in both NMI and average purity. 
 For example, when the missing rate is 0.7, the NMI obtained from CoKL+KCCA is about 0.7, 
 while that of the comparison methods is only about 0.3.
 The average purity obtained from CoKL+KCCA is about 0.85, while that of the comparison methods is only less than 0.65.
 Even when the missing rate is 0.9, the NMI obtained from CoKL+KCCA is still 0.43,
 which is much larger than that of the comparison methods. 
 Of course, the result of proposed algorithm is not as good as the results of algorithms running on complete dataset.
 However, it is important to note that in Fig.~\ref{fig:seeds_2} the proposed algorithm is very closed to the algorithms running on complete dataset in average purity. 
 From Fig.~\ref{fig:seeds2}, it can be easily observed that CoKL+KCCA outperforms CoKL-KA-SC and CoKL-KP-SC almost everywhere, 
 which shows the effectiveness of KCCA.
 These results shows that CoKL+KCCA performs not only better than the intuitive strategy which directly uses the concatenated features, 
 but also better than the latest method MVC.
 
\subsection{Handwritten Dutch Numbers Recognition}
\begin{figure*}[]
\centering
\subfloat[Fourier coefficients and pixel averages] {
\includegraphics[width=0.3\textwidth]{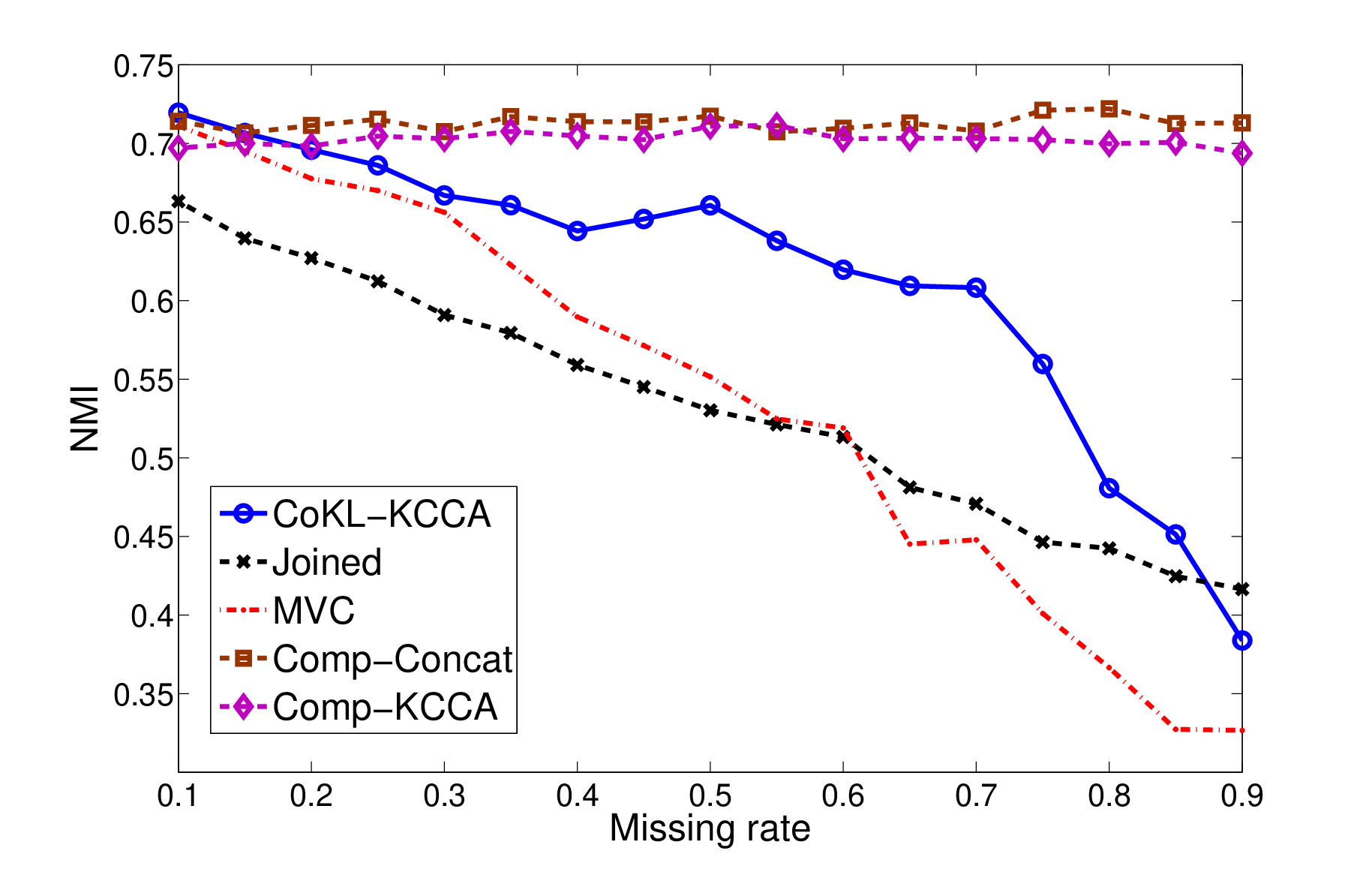}
\label{fig:digit_1}
}
\subfloat[Fourier coefficients and Zernike moments] {
\includegraphics[width=0.3\textwidth]{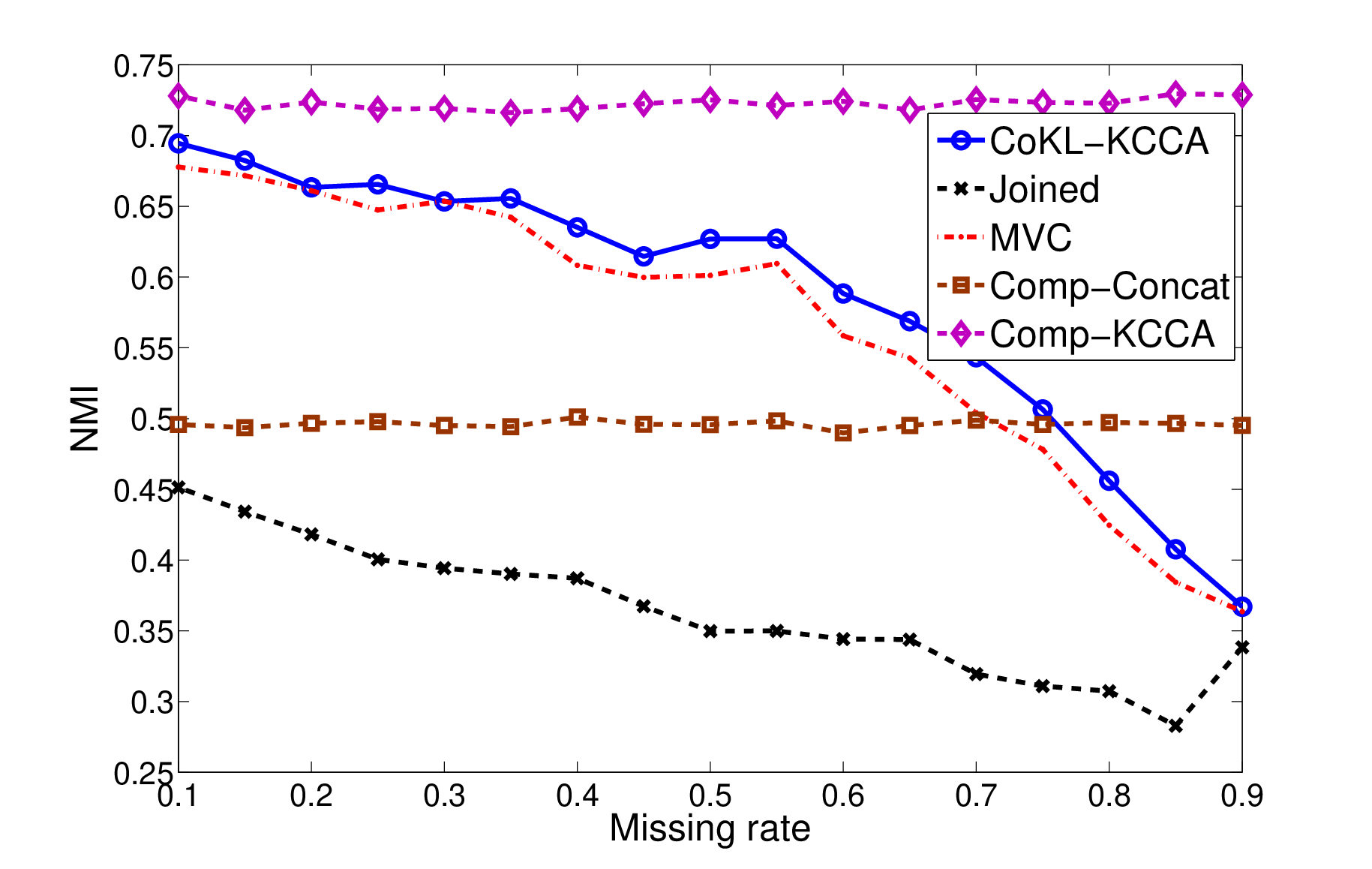}
\label{fig:digit_2}
}
\subfloat[Pixel averages and Zernike moments] {
\includegraphics[width=0.3\textwidth]{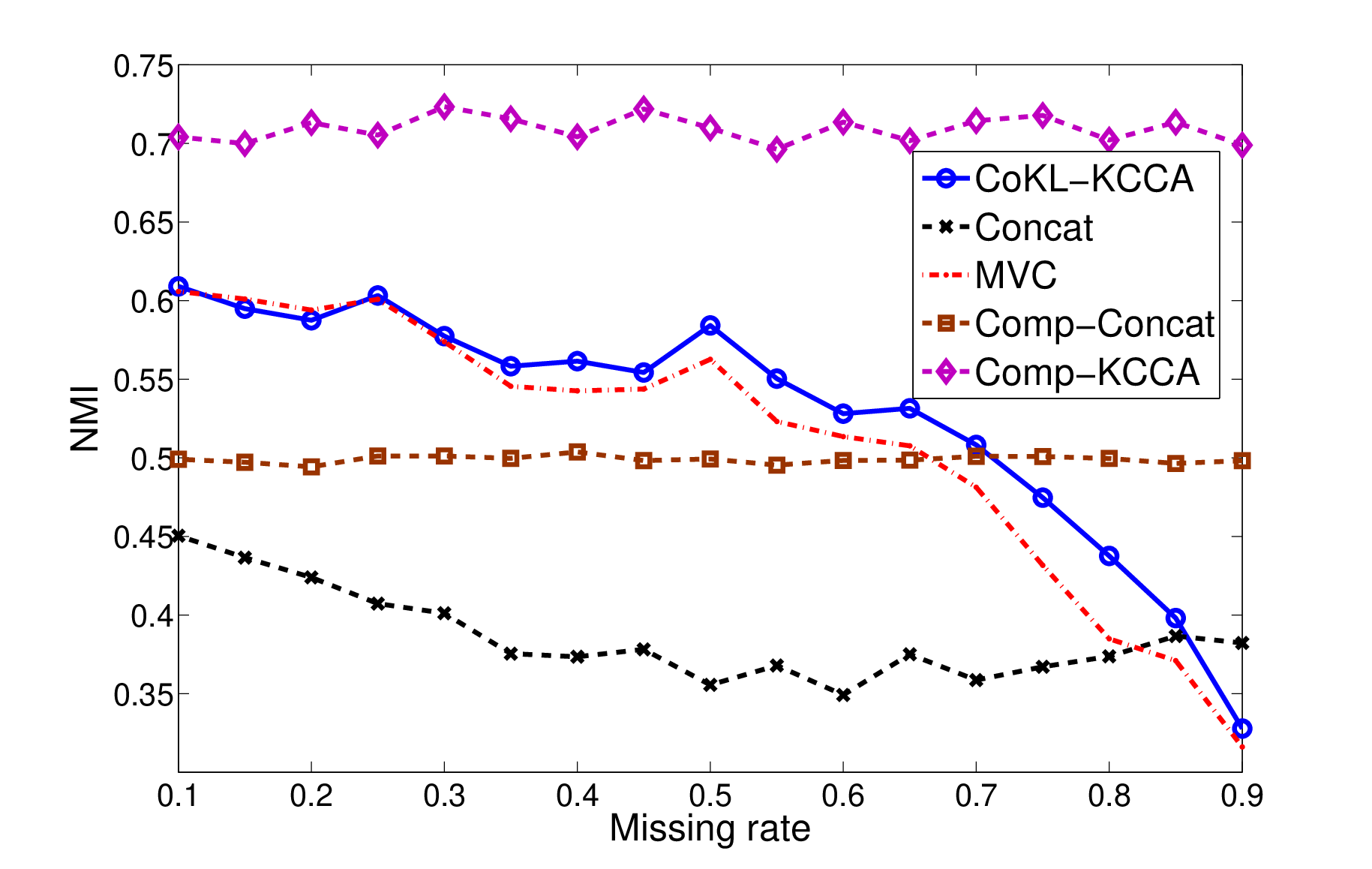}
\label{fig:digit_3}
}\\
\subfloat[Profile correlations and Fourier coefficients] {
\includegraphics[width=0.3\textwidth]{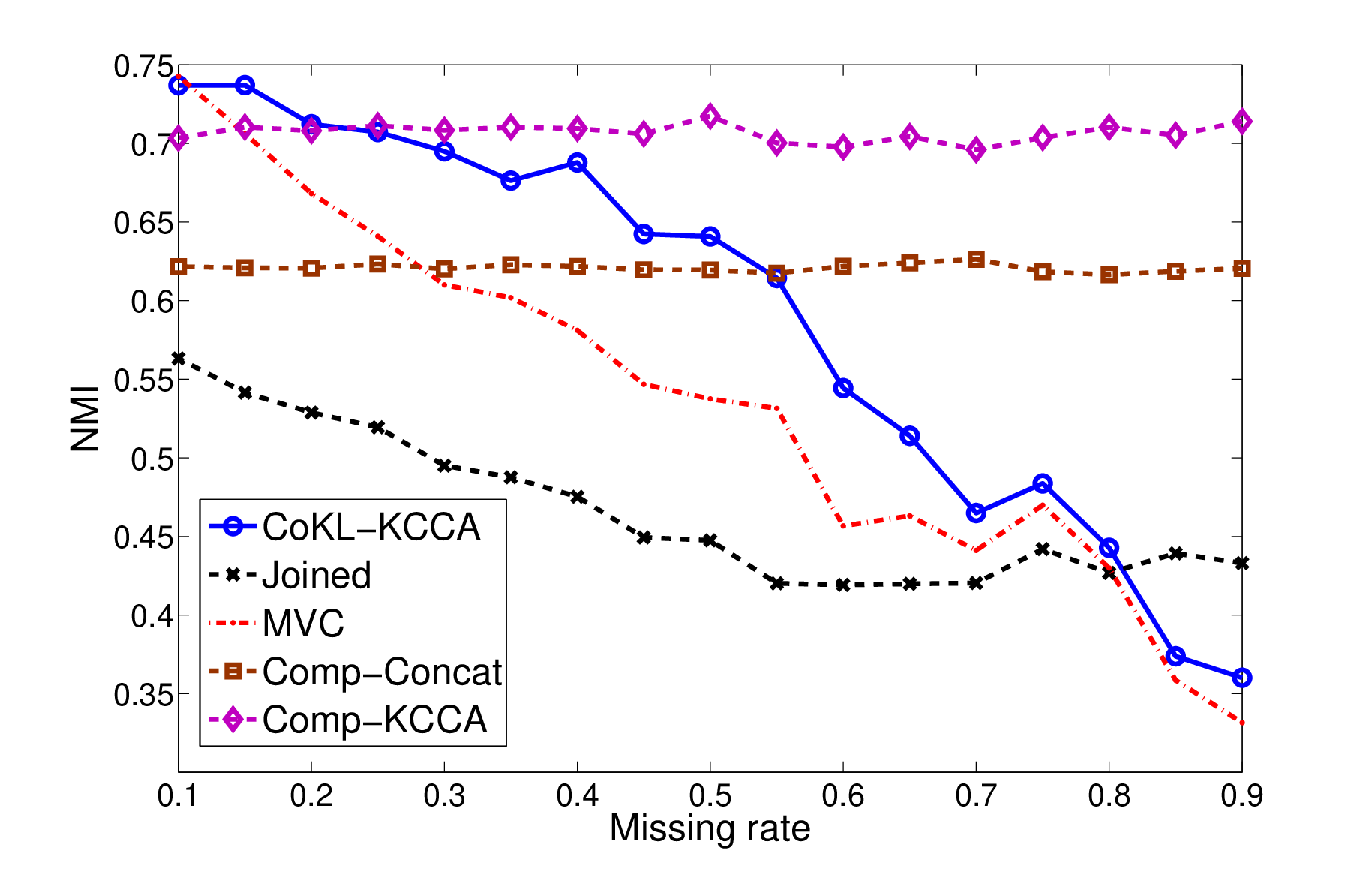}
\label{fig:digit_4}
}
\subfloat[Profile correlations and Zernike moments] {
\includegraphics[width=0.3\textwidth]{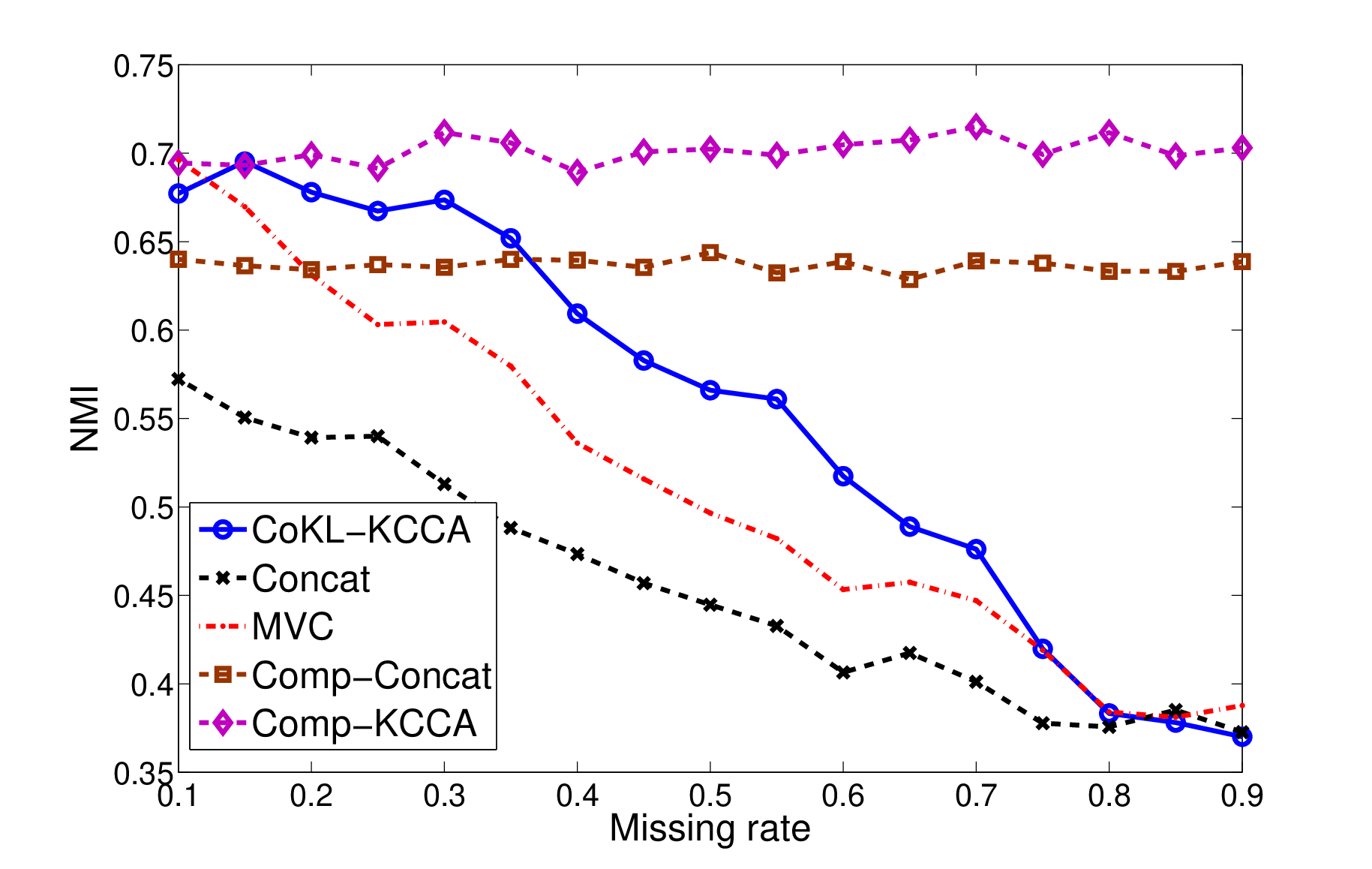}
\label{fig:digit_5}
}
\subfloat[Profile correlations and Pixel averages] {
\includegraphics[width=0.3\textwidth]{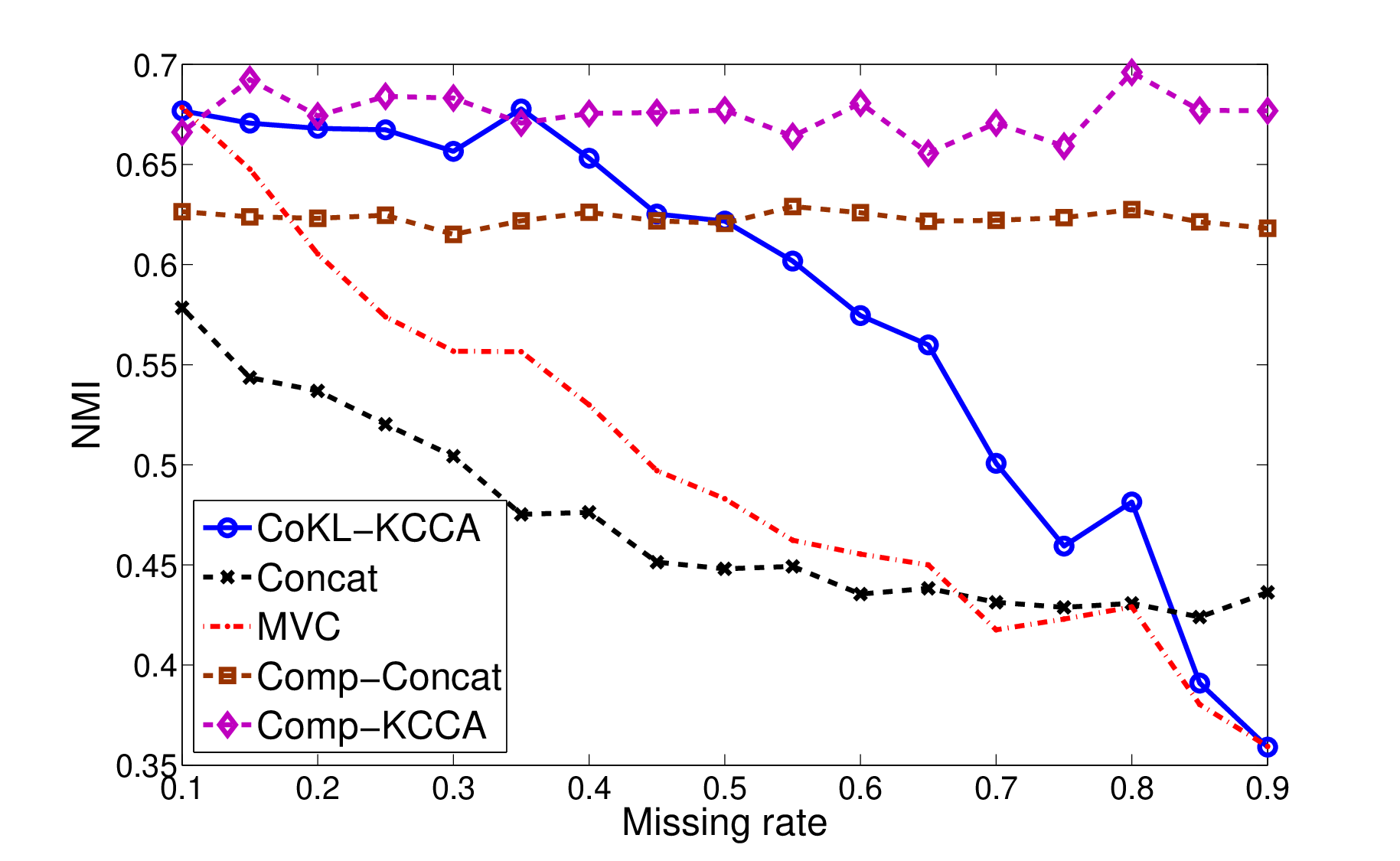}
\label{fig:digit_6}
}\\
\subfloat[Fourier coefficients and Pixel averages] {
\includegraphics[width=0.3\textwidth]{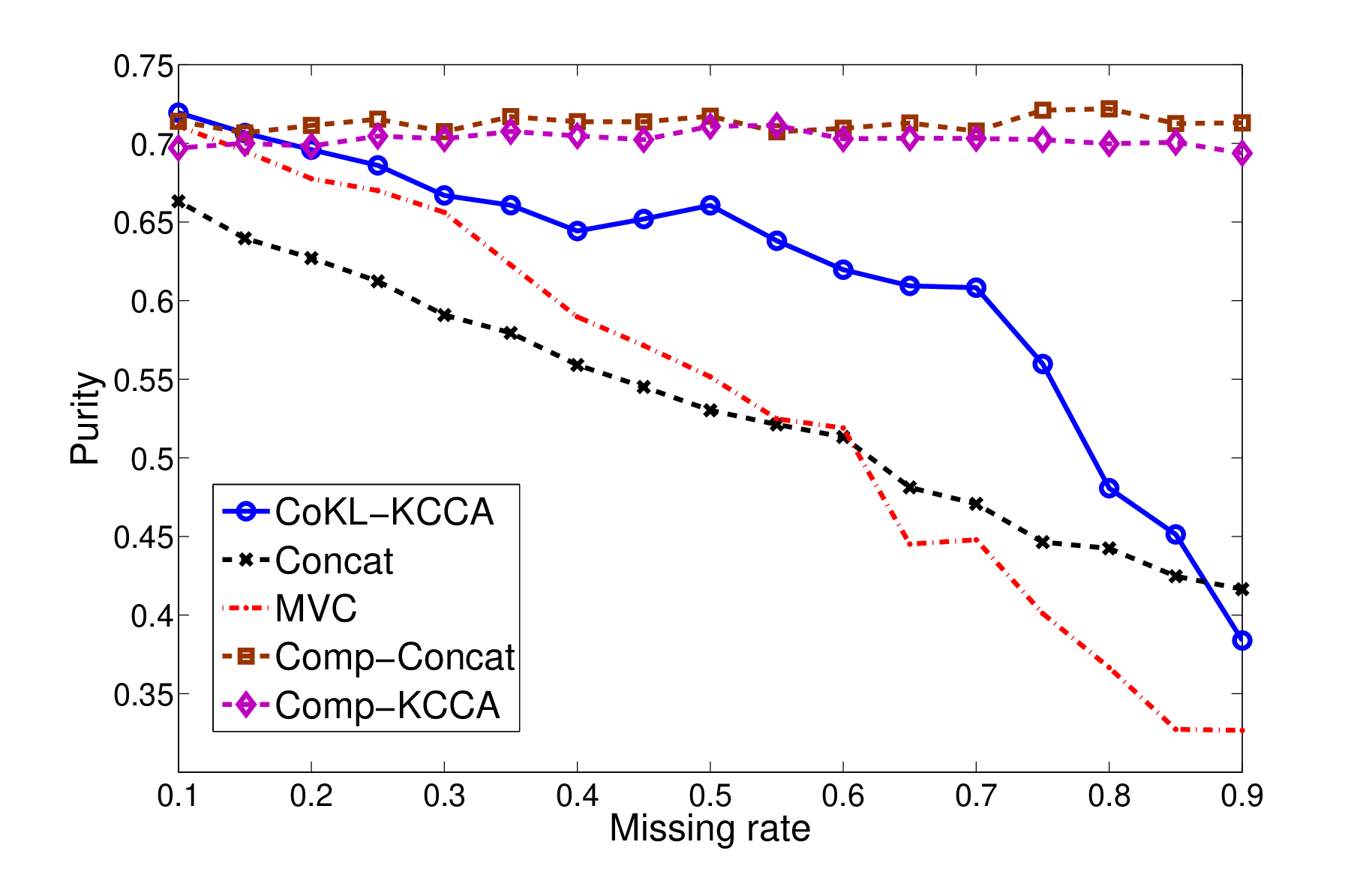}                
\label{fig:digit_1_p}
}
\subfloat[Fourier coefficients and Zernike moments]{
 \includegraphics[width=0.3\textwidth]{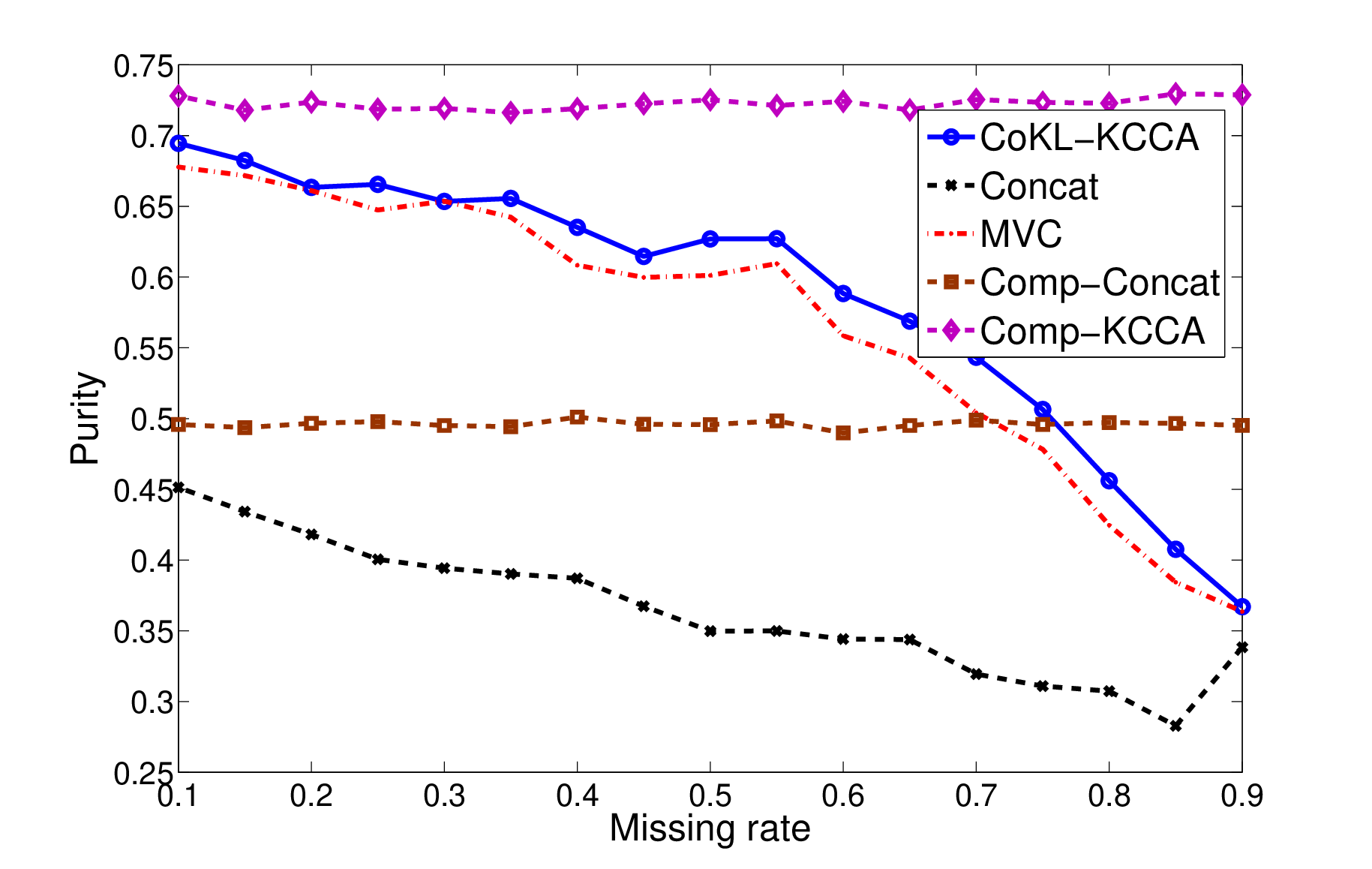}
                \label{fig:digit_2_p}
}
\subfloat[Pixel averages and Zernike moments] {
 \includegraphics[width=0.3\textwidth]{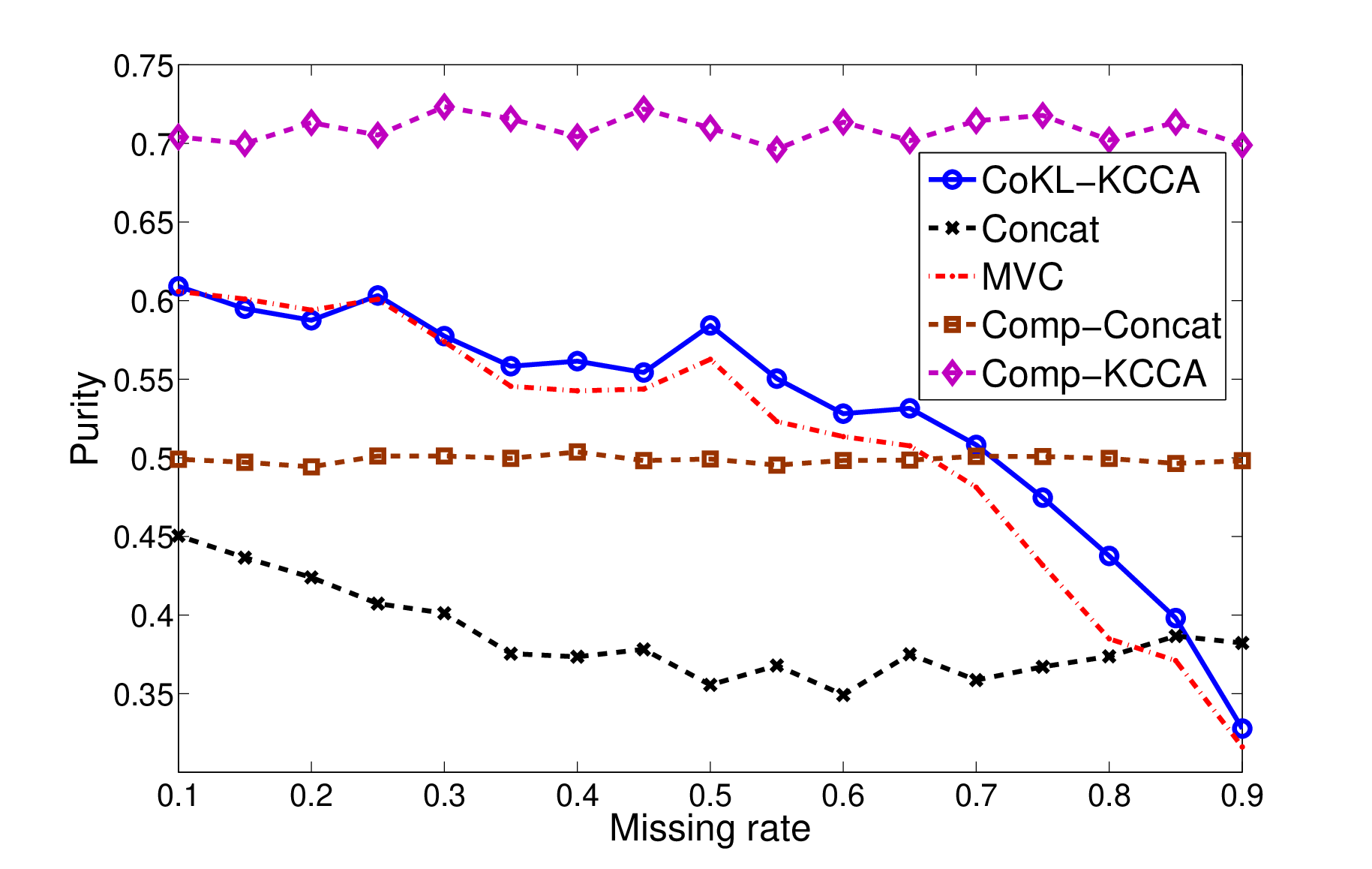}
                \label{fig:digit_3_p}
}\\
\subfloat[Profile correlations and Fourier coefficients] {
\includegraphics[width=0.3\textwidth]{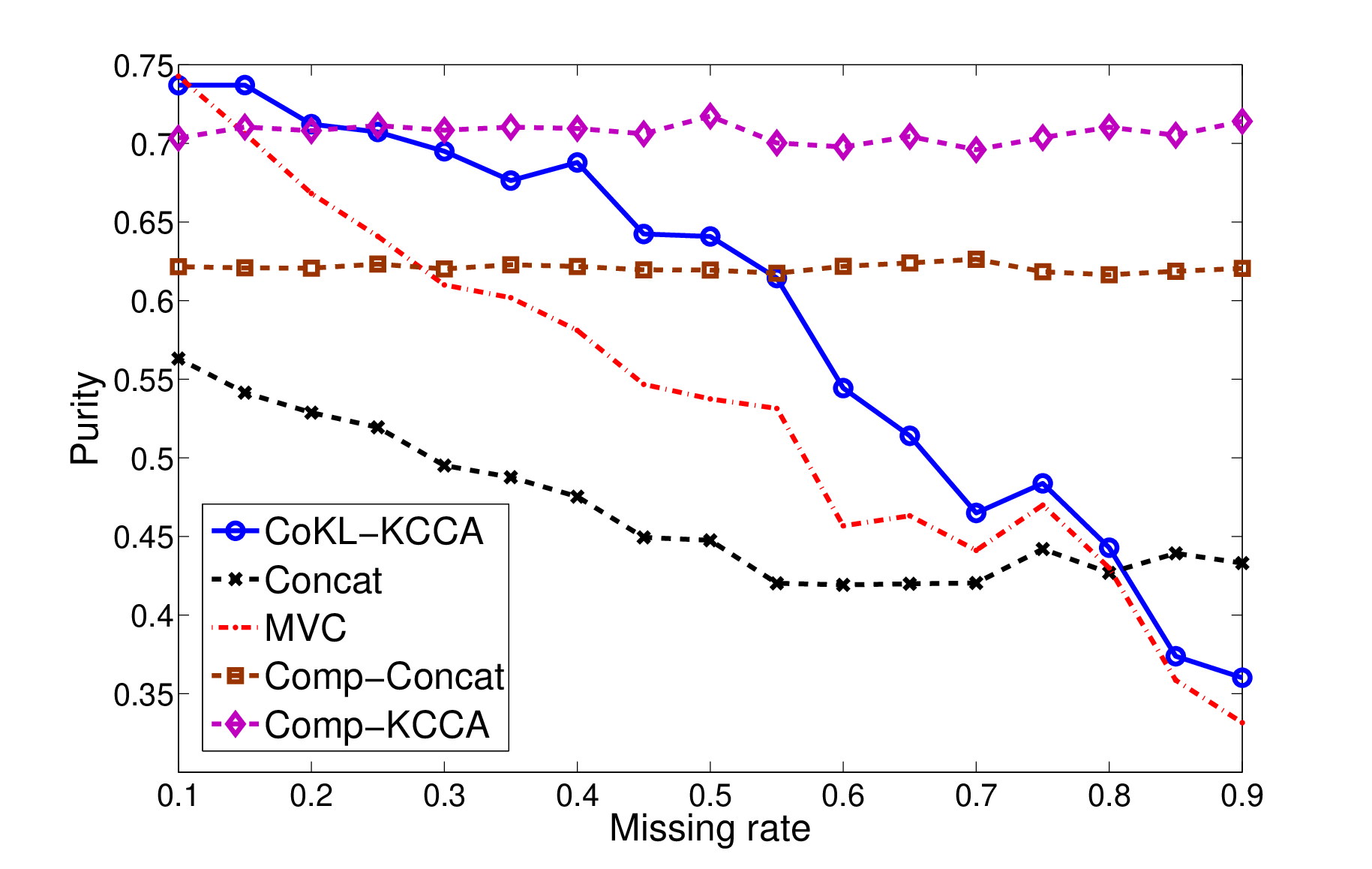}
\label{fig:digit_4_p}
}
\subfloat[Profile correlations and Zernike moments]{
\includegraphics[width=0.3\textwidth]{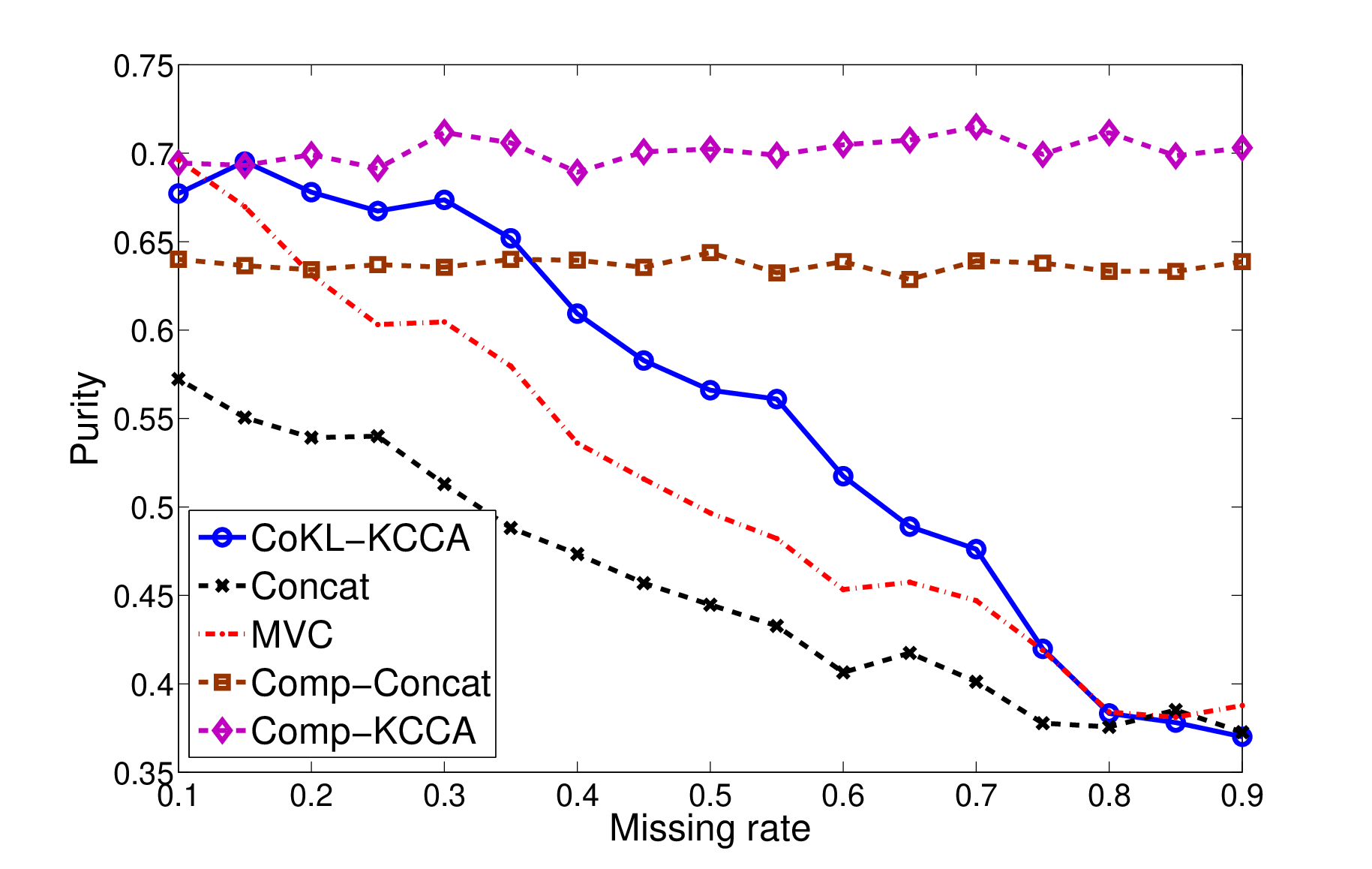}
                \label{fig:digit_5_p}
}
\subfloat[Profile correlations and Pixel averages]{
\includegraphics[width=0.3\textwidth]{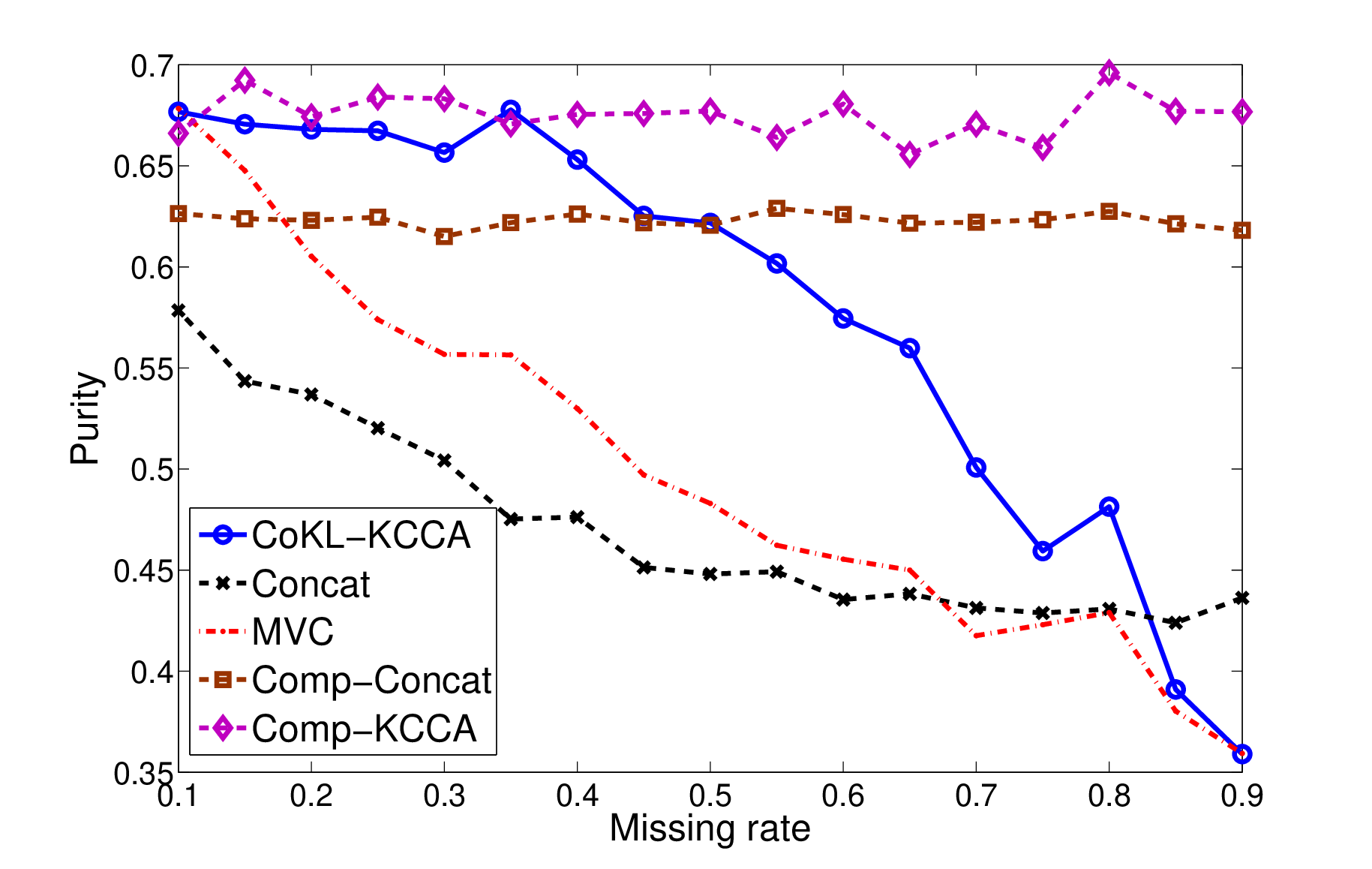}
                \label{fig:digit_6_p}
}
\caption{NMI and average purity on 6 pairs of handwritten Dutch numbers datasets}
\label{fig:digit}
\end{figure*}

\begin{figure*}
\centering
\subfloat[Fourier coefficients and pixel averages] {
\includegraphics[width=0.3\textwidth]{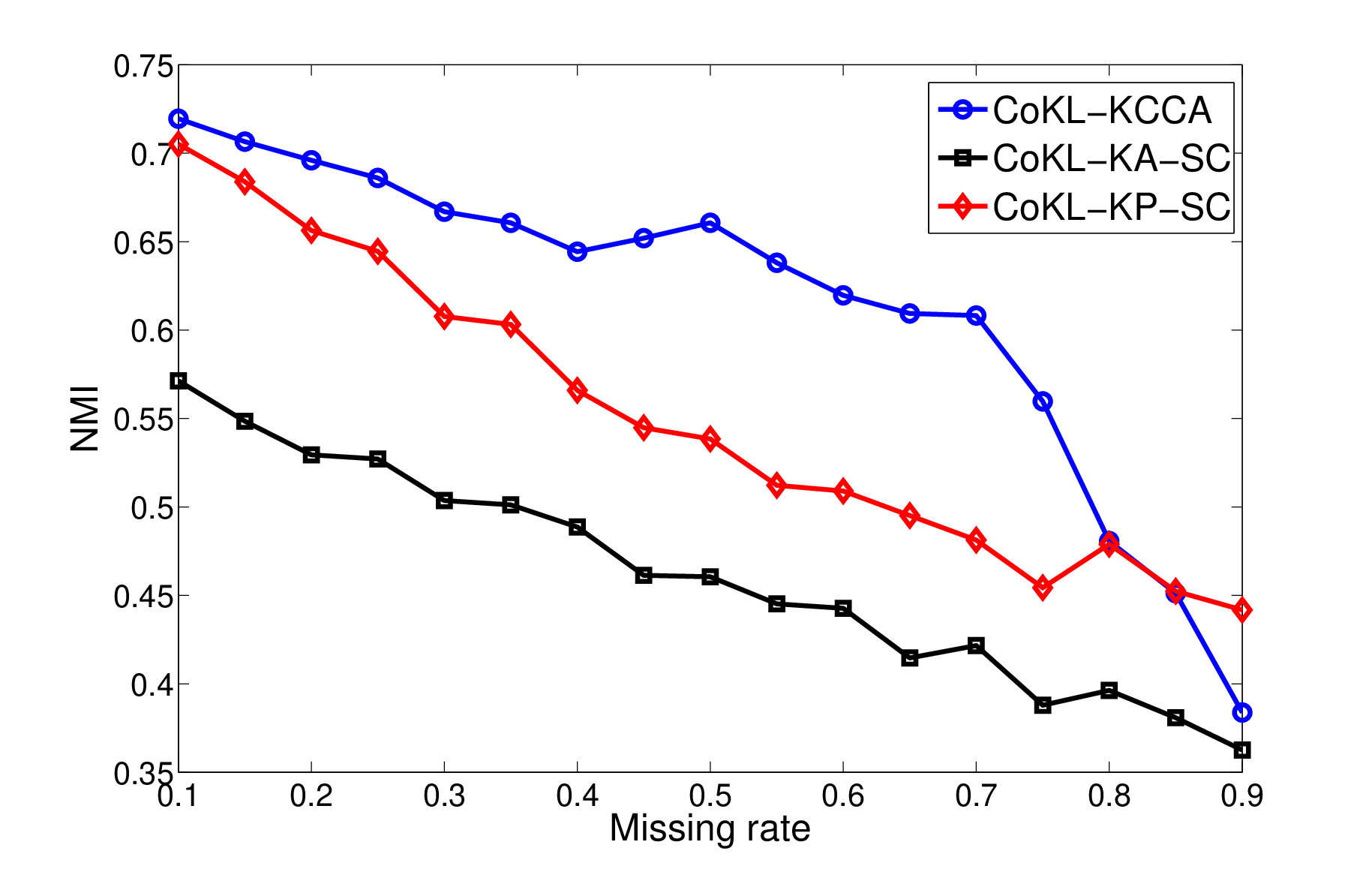}
\label{fig:digit_1_compare}
}
\subfloat[Fourier coefficients and zernike moments] {
\includegraphics[width=0.3\textwidth]{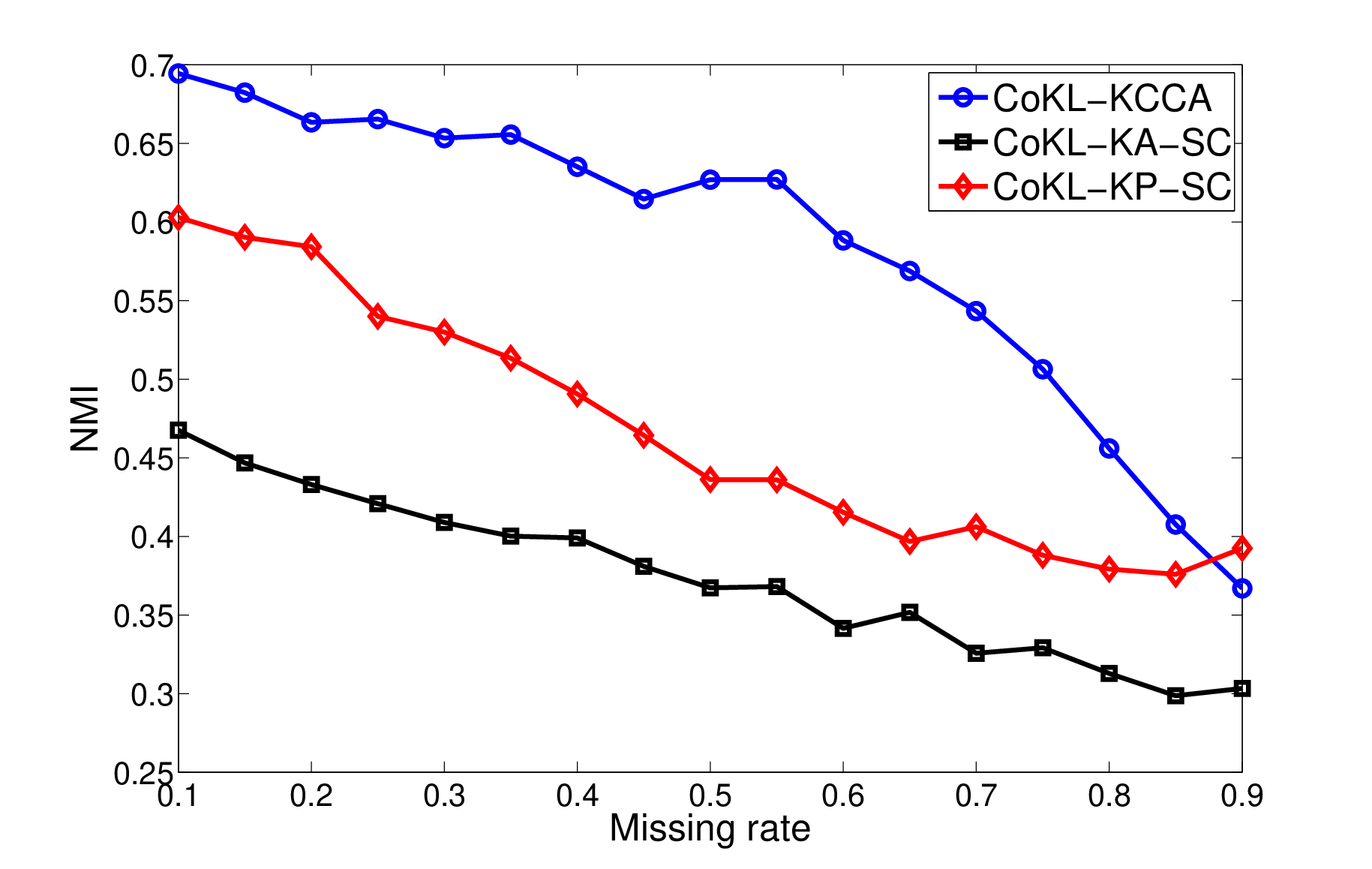}
\label{fig:digit_2_compare}
}
\subfloat[Pixel averages and zernike moments] {
\includegraphics[width=0.3\textwidth]{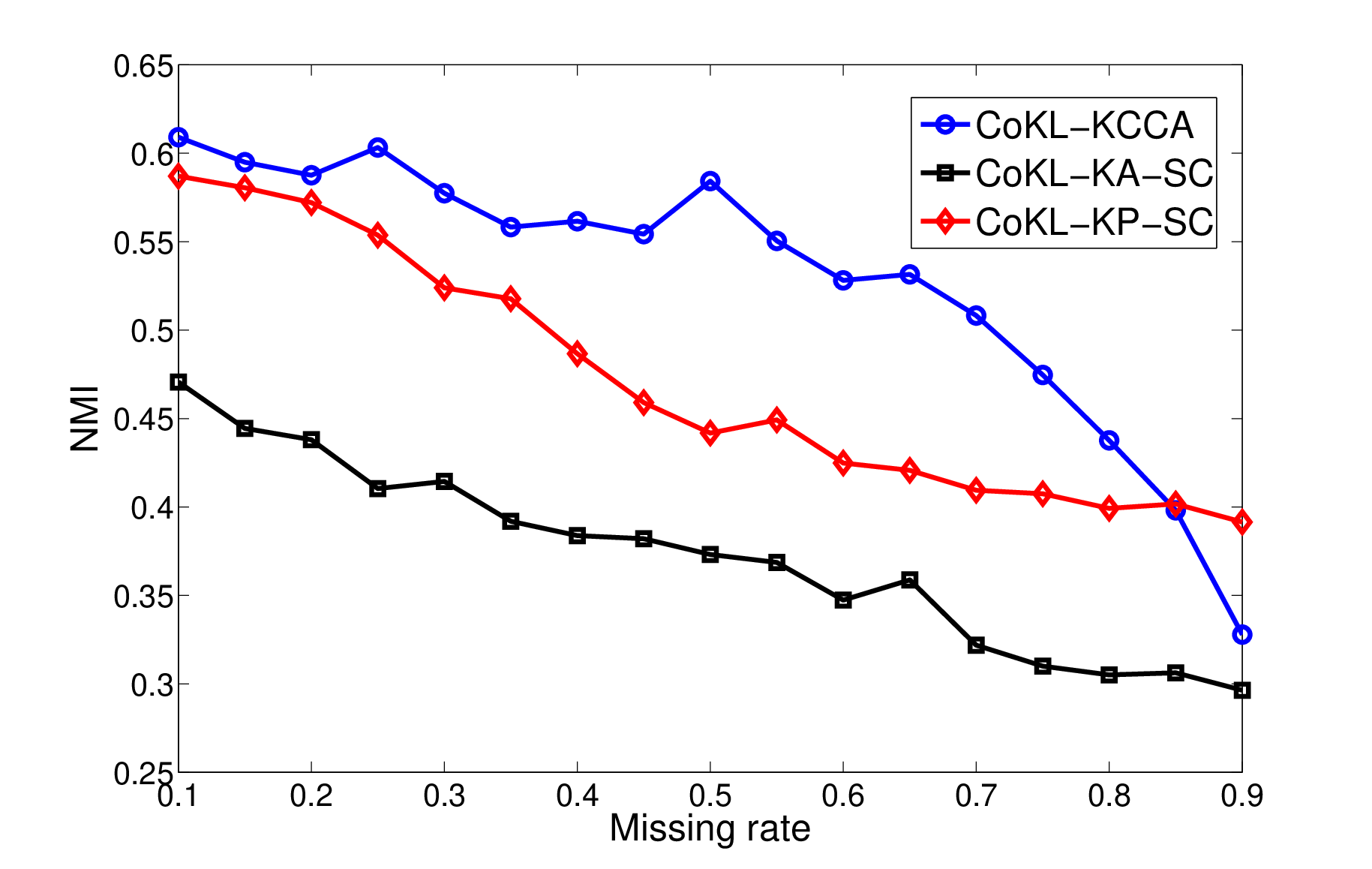}
\label{fig:digit_3_compare}
}\\
\subfloat[Profile correlations and fourier coefficients] {
\includegraphics[width=0.3\textwidth]{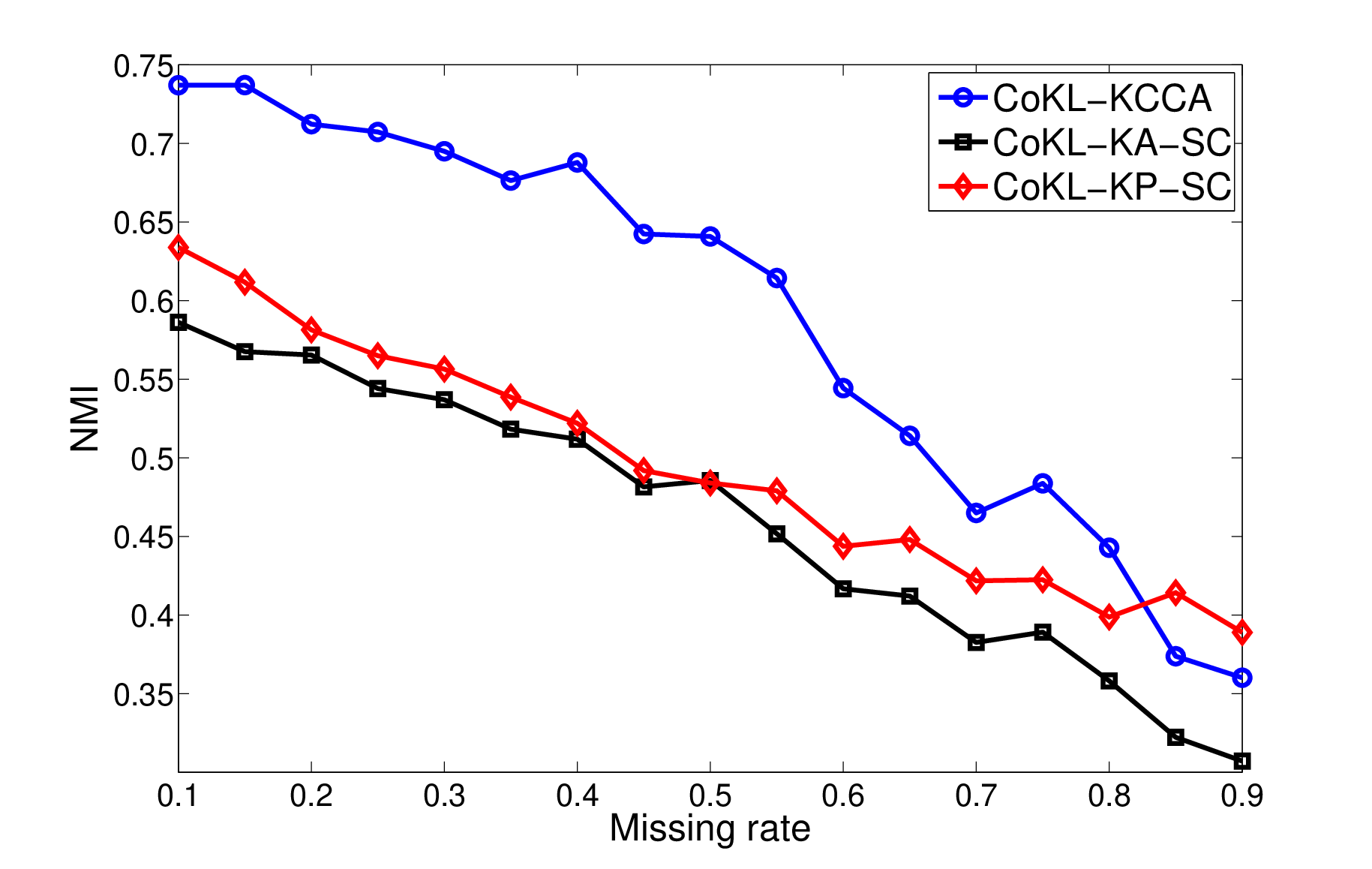}
\label{fig:digit_4_compare}
}
\subfloat[Profile correlations and zernike moments] {
\includegraphics[width=0.3\textwidth]{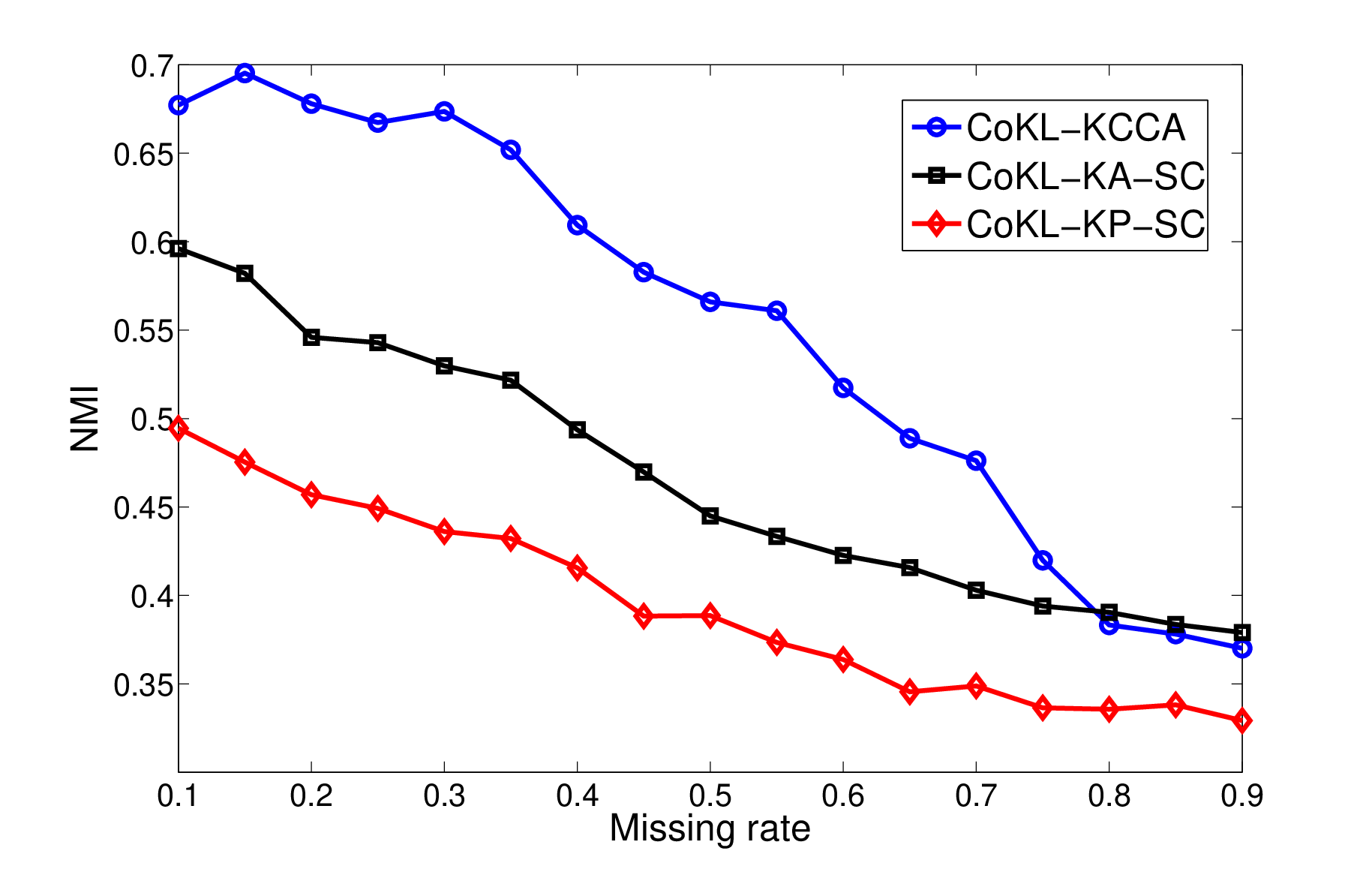}
\label{fig:digit_5_compare}
}
\subfloat[Profile correlations and pixel averages] {
\includegraphics[width=0.3\textwidth]{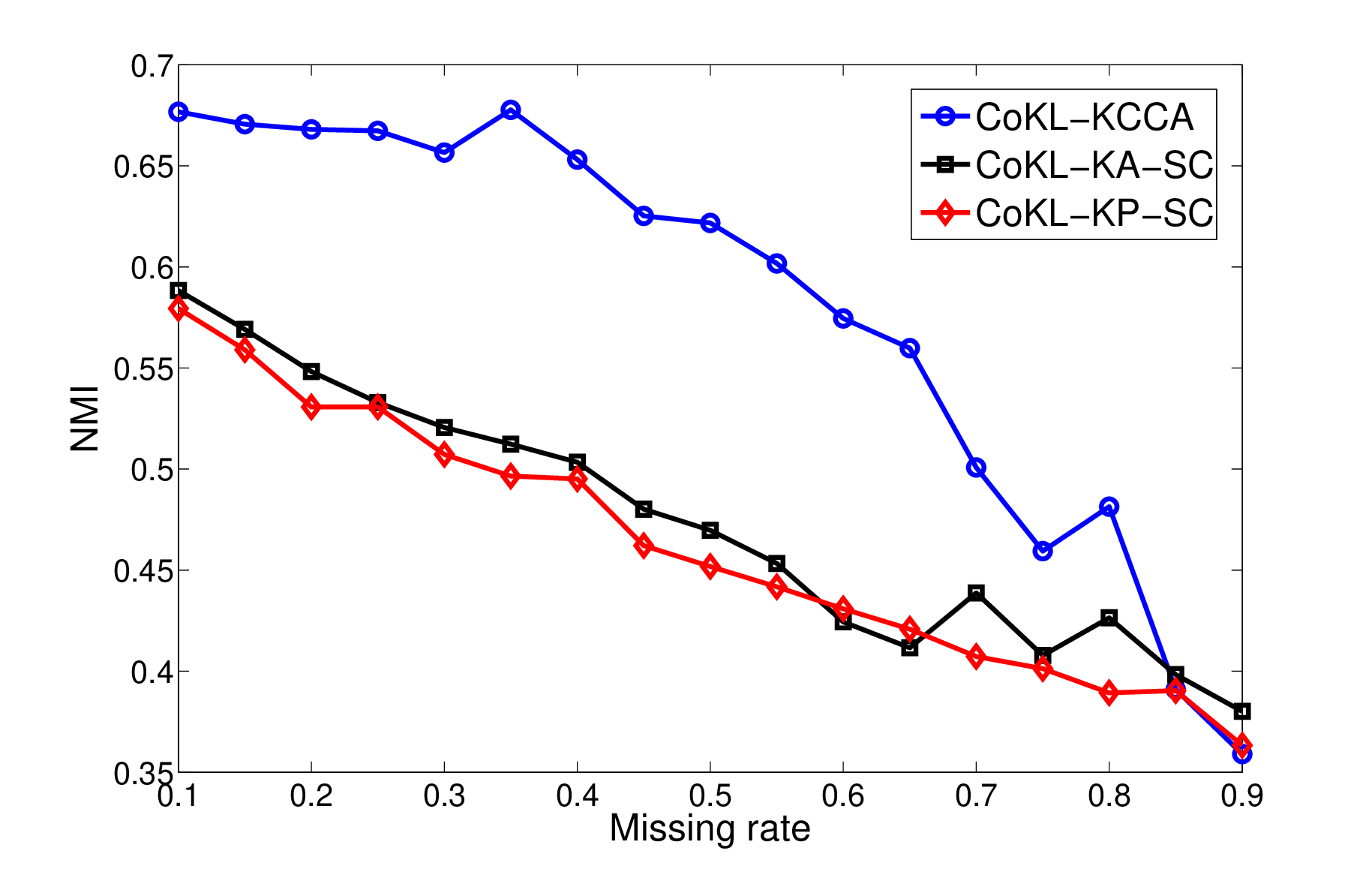}
\label{fig:digit_6_compare}
}\\
\subfloat[Fourier coefficients and pixel averages] {
\includegraphics[width=0.3\textwidth]{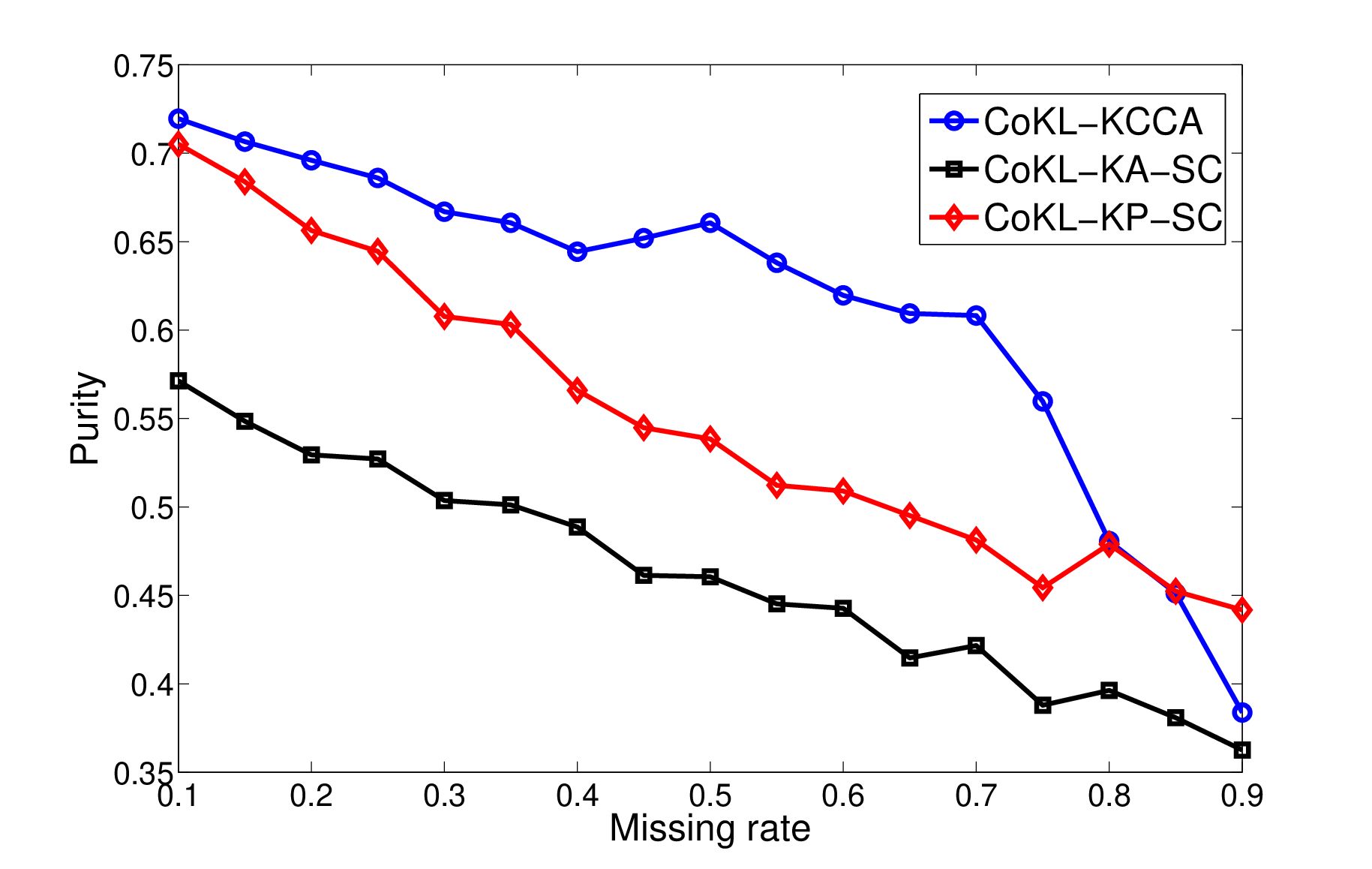}                
\label{fig:digit_1_p_compare}
}
\subfloat[Fourier coefficients and zernike moments]{
 \includegraphics[width=0.3\textwidth]{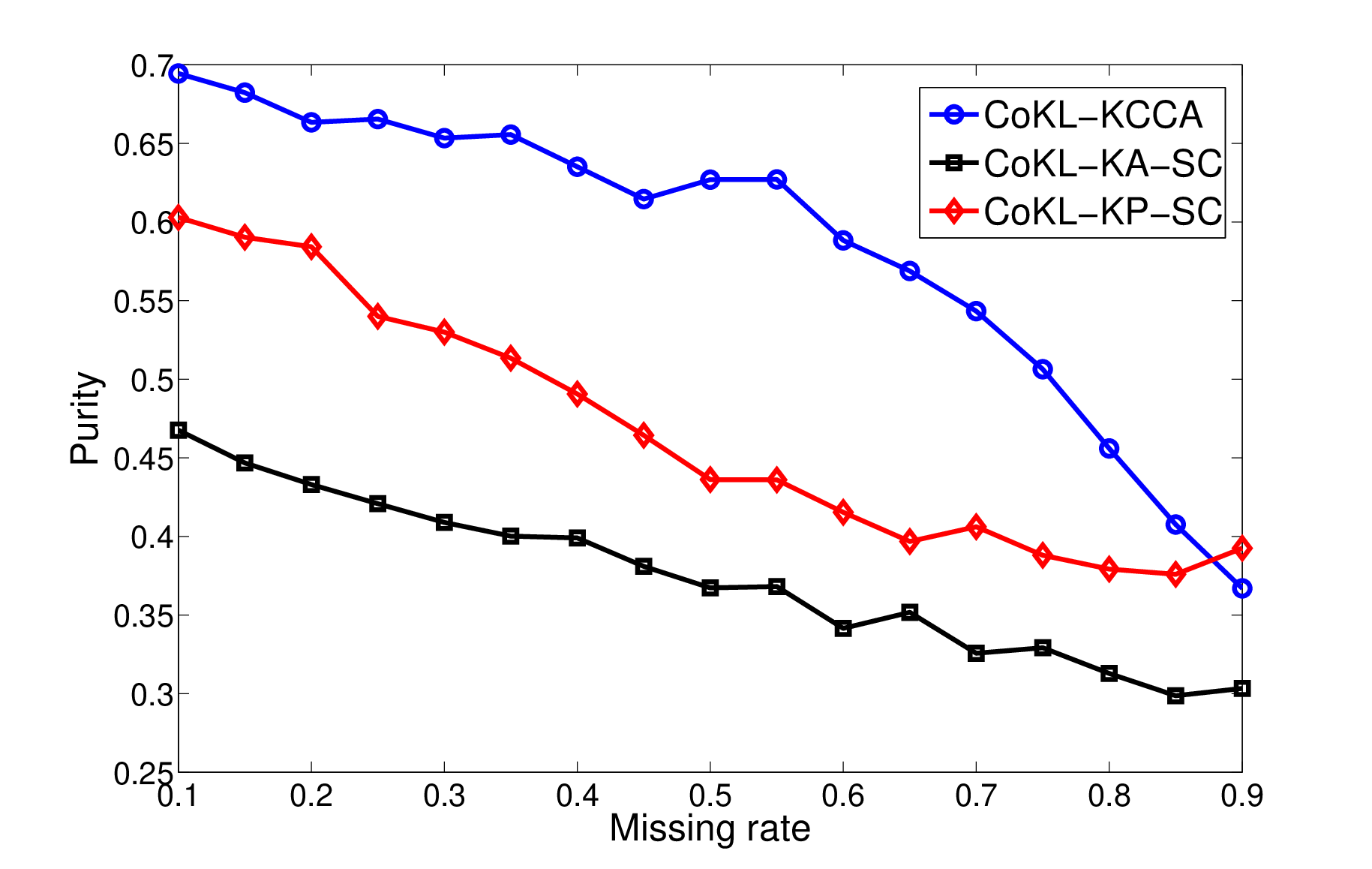}
                \label{fig:digit_2_p_compare}
}
\subfloat[Pixel averages and zernike moments] {
 \includegraphics[width=0.3\textwidth]{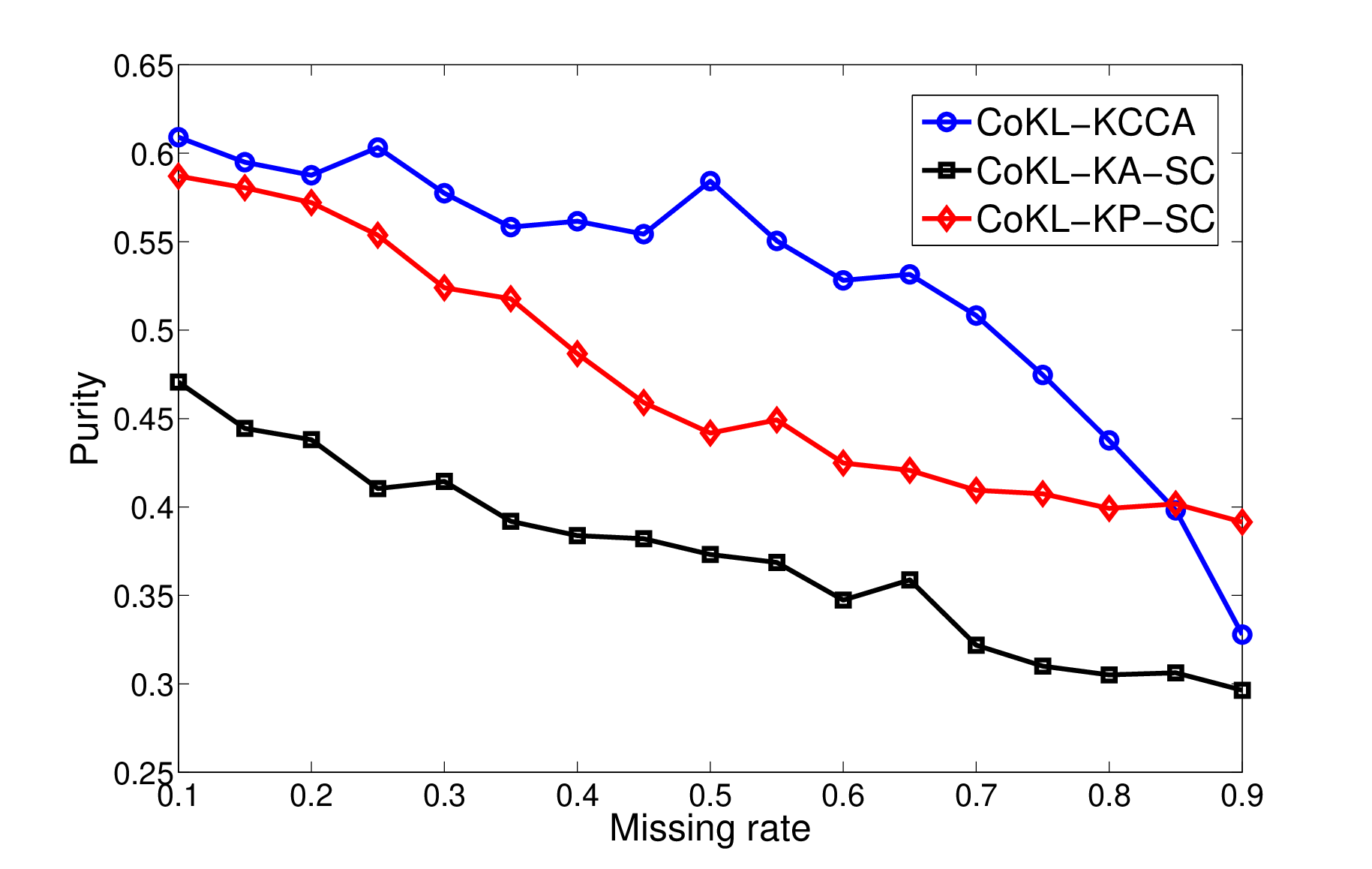}
                \label{fig:digit_3_p_compare}
}\\
\subfloat[Profile correlations and fourier coefficients] {
\includegraphics[width=0.3\textwidth]{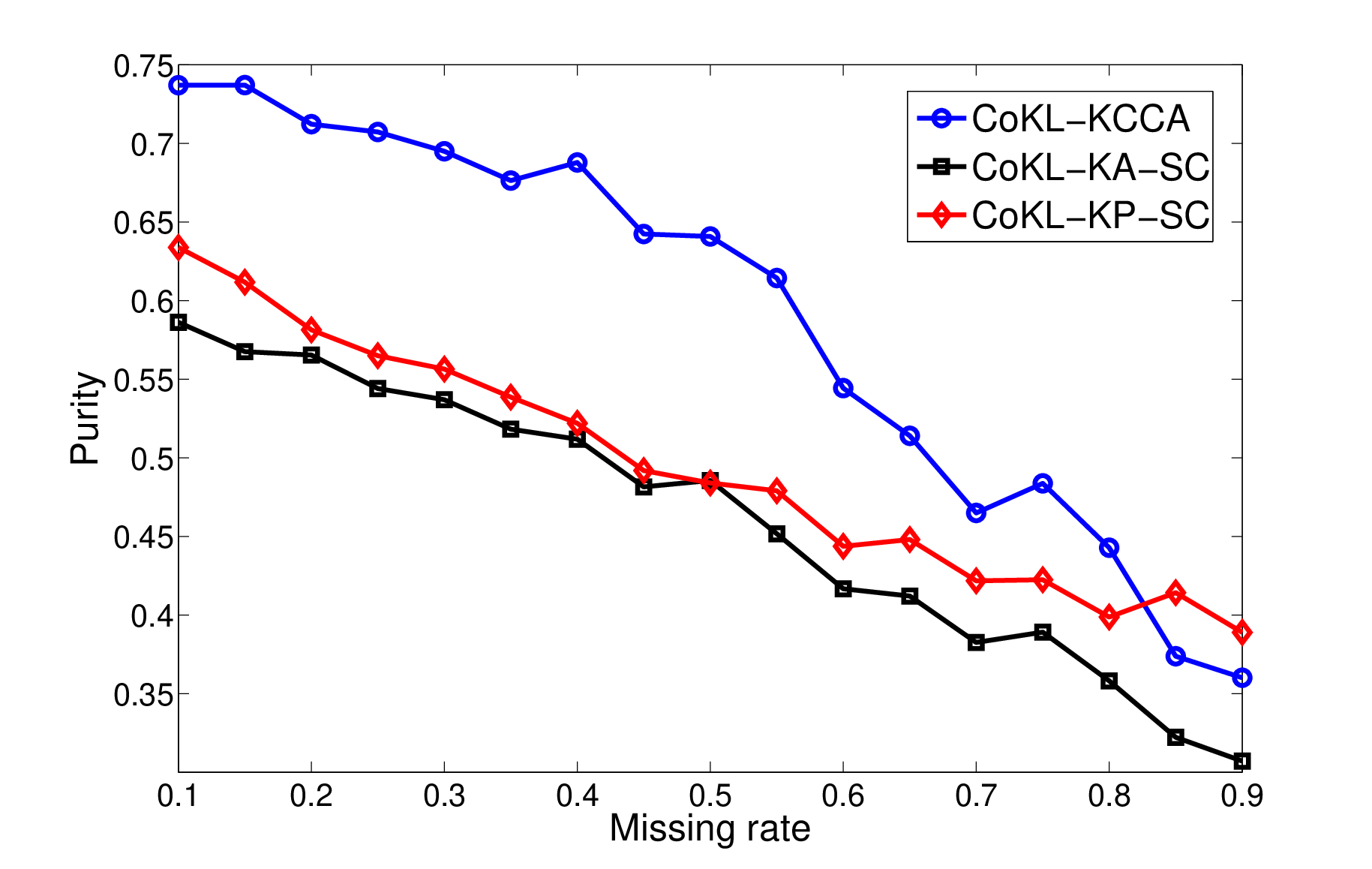}
\label{fig:digit_4_p_compare}
}
\subfloat[Profile correlations and zernike moments]{
\includegraphics[width=0.3\textwidth]{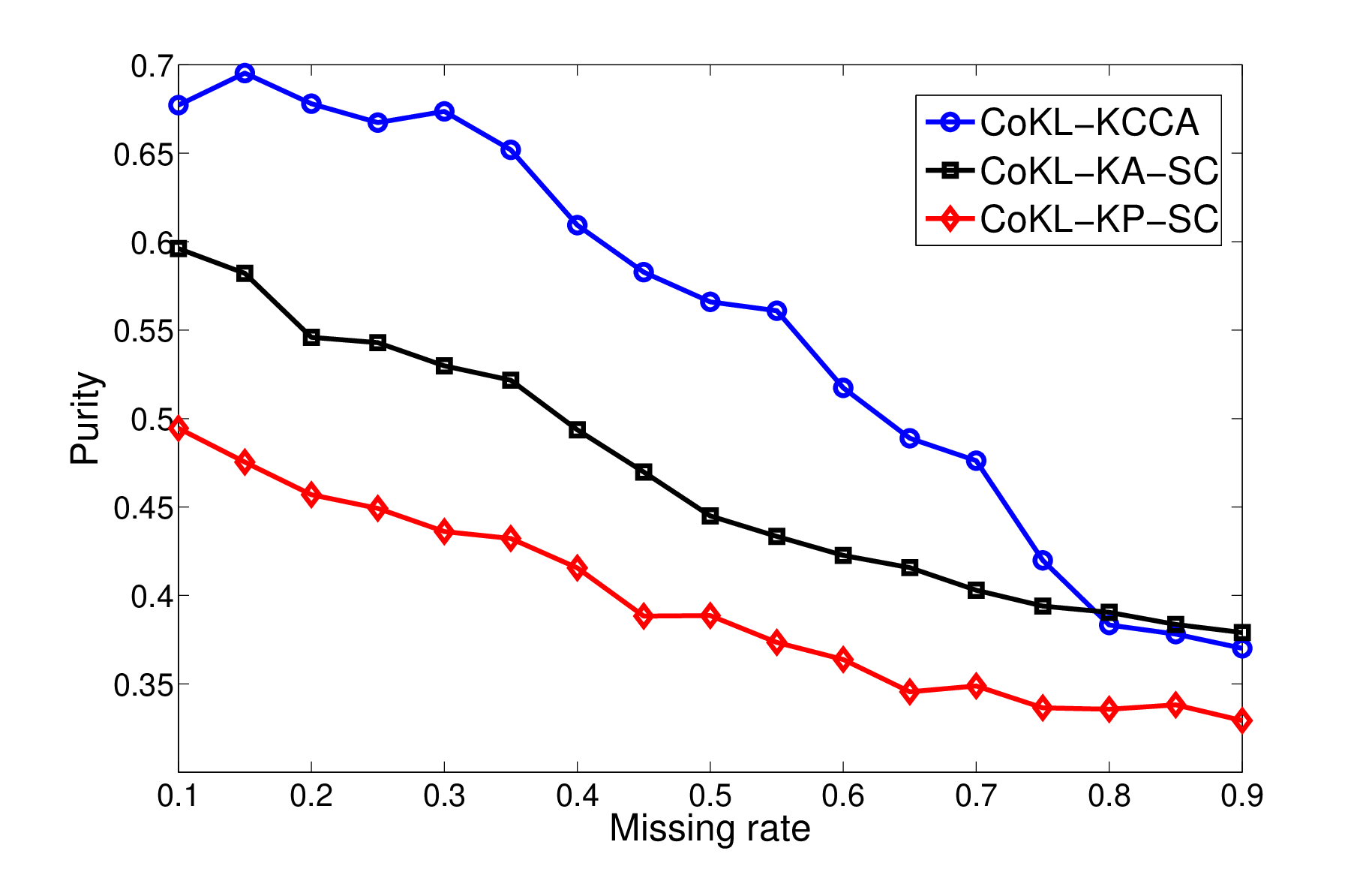}
                \label{fig:digit_5_p_compare}
}
\subfloat[Profile correlations and pixel averages]{
\includegraphics[width=0.3\textwidth]{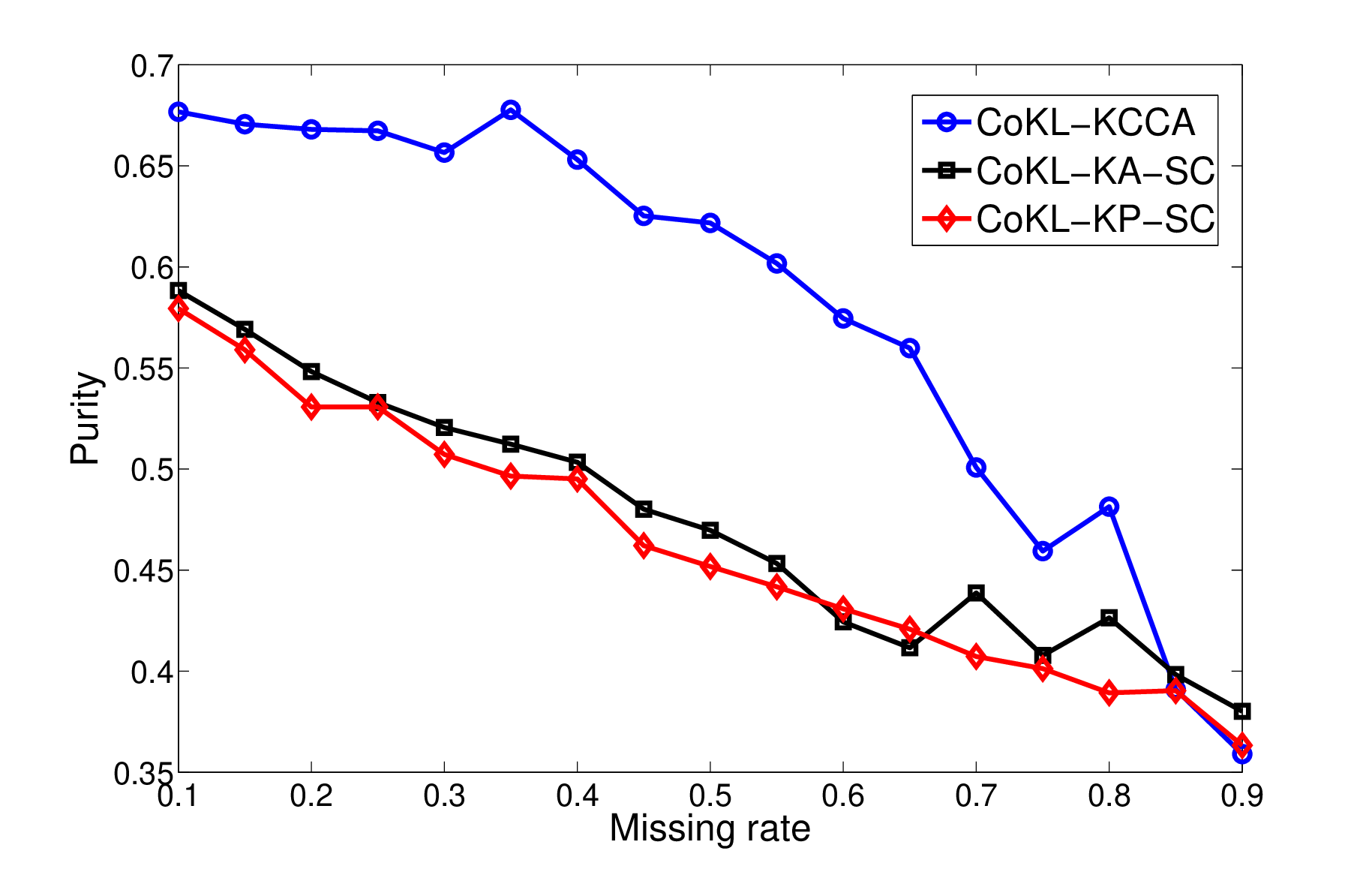}
                \label{fig:digit_6_p_compare}
}                
\caption{The performance of different algorithms combined with CoKL on 6 pairs of handwritten Dutch numbers datasets}
\label{fig:digit_purity}
\end{figure*}

This dataset contains 2000 handwritten numerals ("0"-"9") extracted from a collection of Dutch utility maps \cite{dataset:digit}. 
The handwritten numbers are scanned and digitized as binary images. 
The following feature spaces (datasets) with different vector-based features is available for the numbers:
(1) 76 Fourier coefficients of the character shapes, (2) 216 profile correlations, (3) 240 pixel averages in $2\times3$ windows,
and (4) 47 Zernike moments. All these features are conventional vector-based features but in different feature spaces. 
The aim is to cluster the numbers. 
We test the proposed algorithm on two incomplete datasets, so among this 4 different datasets, we can have 6 different combinations.
For each pair of datasets, we randomly delete the instances in both of the datasets. 
As mentioned before, we run the proposed algorithm for different total missing rates (from 10\% to 90\%). 
The results of all the 6 different combinations average over 10 runs are presented in Fig.~\ref{fig:digit} and Fig.~\ref{fig:digit_purity}. 

%\begin{figure*}
%\centering
%\subfloat[Fourier coefficients and pixel averages] {
%\includegraphics[width=0.3\textwidth]{../experiment_plot/nmi/f_pi_CoKL_compare_nmi.eps}
%\label{fig:digit_1_compare}
%}
%\subfloat[Fourier coefficients and Zernike moments] {
%\includegraphics[width=0.3\textwidth]{../experiment_plot/nmi/f_z_CoKL_compare_nmi.eps}
%\label{fig:digit_2_compare}
%}
%\subfloat[Pixel averages and Zernike moments] {
%\includegraphics[width=0.3\textwidth]{../experiment_plot/nmi/pi_z_CoKL_compare_nmi.eps}
%\label{fig:digit_3_compare}
%}\\
%\subfloat[Profile correlations and Fourier coefficients] {
%\includegraphics[width=0.3\textwidth]{../experiment_plot/nmi/p_f_CoKL_compare_nmi.eps}
%\label{fig:digit_4_compare}
%}
%\subfloat[Profile correlations and Zernike moments] {
%\includegraphics[width=0.3\textwidth]{../experiment_plot/nmi/p_z_CoKL_compare_nmi.eps}
%\label{fig:digit_5_compare}
%}
%\subfloat[Profile correlations and pixel averages] {
%\includegraphics[width=0.3\textwidth]{../experiment_plot/nmi/p_pi_CoKL_compare_nmi.eps}
%\label{fig:digit_6_compare}
%}
%\caption{NMIs of 6 pairs of handwritten Dutch numbers datasets}
%\label{fig:digit_compare}
%\end{figure*}

Fig.~\ref{fig:digit} compares CoKL+KCCA with Concat, MVC, and two algorithms on complete data (Comp-Concat and Comp-KCCA).
Fig.~\ref{fig:digit_purity} compares different algorithms combined with CoKL on incomplete data (CoKL+KCCA, CoKL-KA-SC and CoKL-KP-SC).
As it can be observed, the proposed algorithm, clustering with CoKL and KCCA, 
outperforms the two comparison methods (Concat and MVC) substantially for all the six pairs of datasets in both NMI and average purity.
Taking the pair Fourier coefficients dataset and pixel average dataset (Fig.~\ref{fig:digit_1} and Fig.~\ref{fig:digit_1_p}) as example, 
the NMI obtained from CoKL+KCCA is 0.63 at missing rate 0.7, while that of the comparison methods is only about 0.45. 
The purity obtained from CoKL+KCCA is about 0.7 at missing rate 0.7, while that of the comparison methods is less than 0.5. 
One interesting result is that CoKL+KCCA is even better than Comp-Concat for some settings like Fig.~\ref{fig:digit_2} and Fig~\ref{fig:digit_3}.
The reason is because even with complete dataset, the correlations among the two dataset may not be significant. 
However, the correlations among the
projected spaces between the two sets is maximized when apply KCCA. 
So CoKL+KCCA could be better than Comp-Concat for some settings but is worse than Comp-KCCA for almost every setting.
 From Fig.~\ref{fig:digit_1_p_compare}-\ref{fig:digit_6_p_compare}, it can be easily observed that CoKL+KCCA outperforms CoKL-KA-SC and CoKL-KP-SC for most of the cases, 
 which shows the effectiveness of KCCA.
These results shows that on incomplete datasets, CoKL+KCCA performs not only better than the intuitive strategy Concat and advanced method MVC, 
but even better than some simple algorithms on complete datasets.
\subsection{Discussion}
In this section, we aim at analyzing CoKL more in detail in order to answer the following three questions:
\begin{enumerate}
\item Can we find any patterns and properties of CoKL compared with other approach?
%\item Why is CoKL+KCCA better than CoKL-KA-SC and CoKL-KP-SC?
\item How does the result of clustering using CoKL+KCCA look like geometrically?
\item How does the missing rate affect the convergence rate?
%\item How does CoKL work for more than two dataset?
\end{enumerate}
From Fig.~\ref{fig:seeds} and Fig.~\ref{fig:digit}, we can find the following patterns/properties.
First, for a small missing rate like $10\%$, 
The performance of CoKL+KCCA and MVC are almost the same, but both better than the intuitive strategy Concat.
Second, as missing rate goes larger, the performance of all these three methods decline. 
The performance of CoKL+KCCA drops slower than MVC.
Third, when the missing rate is 0.9, the performance of all these three methods get to the lowest point. 
CoKL+KCCA and MVC may be worse than the intuitive strategy which directly uses the concatenated features.

These patterns and properties make sense. 
At small missing rate, there is only small amount of instances missing in datasets.
The difference between CoKL+KCCA and MVC is really small. 
As the missing rate goes larger, the information contained in the missing instances become more and more. 
The information that all these three methods could obtain become less and less. 
That would explain the performance drop for all these three methods.
MVC assumes one dataset is complete and only completes the other kernel matrix once, while CoKL takes advantage of the common examples in two datasets and collectively complete the kernel matrices of the datasets. So CoKL+KCCA can get more information from the incomplete datasets.
Thus, CoKL+KCCA performs better than MVC.
However, when the missing rate is too large, say 0.9, the portion of useful information in the initial kernel matrices (filling the datasets with average/majority values) is so small and biased that CoKL and MVC may be misled by the small portion of examples. 
Thus, the performance of CoKL+KCCA and MVC may be slightly worse than the intuitive strategy at a large missing rate.
\begin{figure}
\includegraphics[width=\columnwidth]{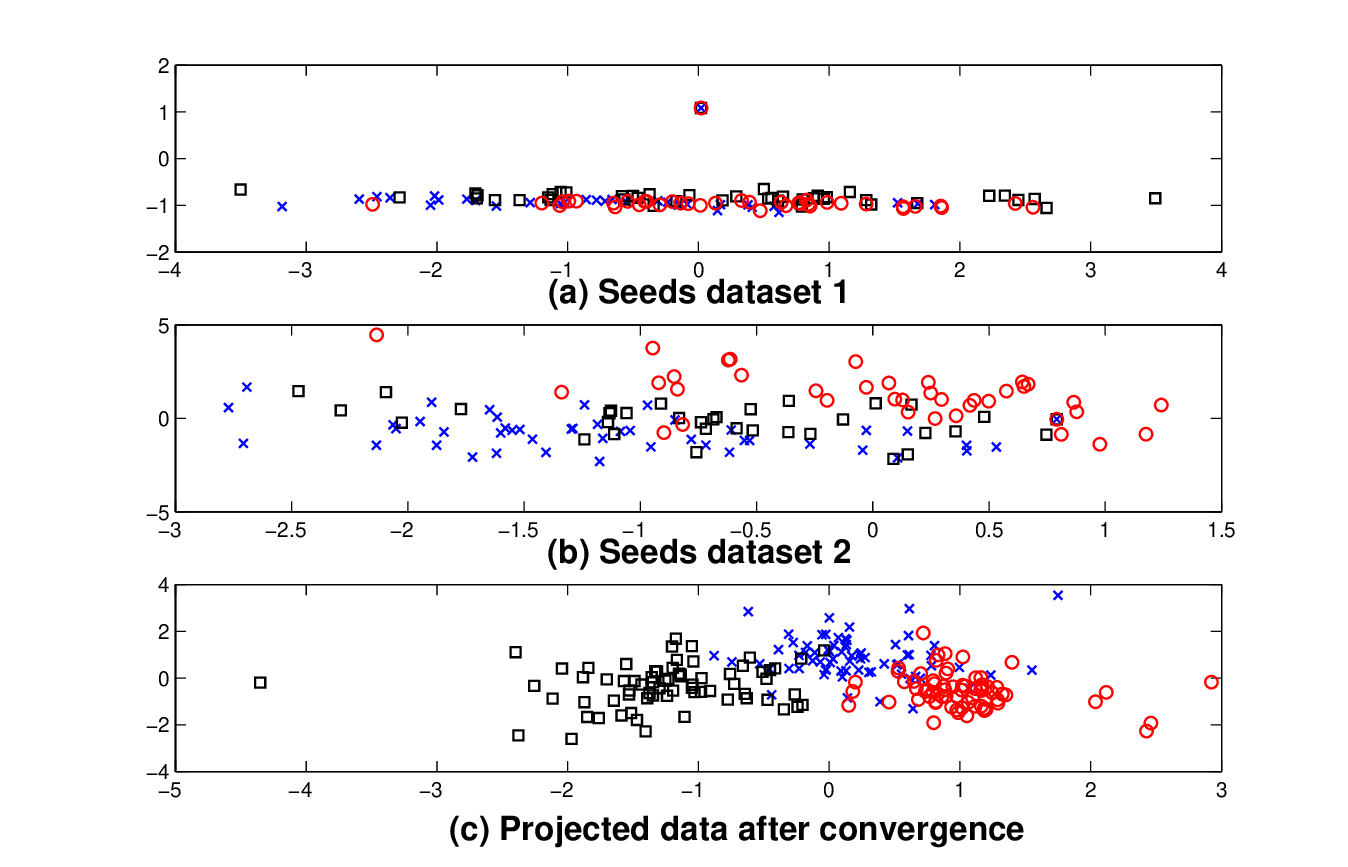}
\caption{The projected data: $\Box$, x, o are three different classes. }
\label{fig:project}
\end{figure}
%
%From  the results in Section\ref{sec:experiments}, we can find CoKL+KCCA outperforms CoKL-KA-SC and CoKL-KP-SC almost everywhere.
%As suggested in \cite{NIPS2009_0716} and \cite{NIPS2011_0817}, 

In order to compare the data point before and after CoKL+KCCA clustering, 
we project the UCI seeds data before CoKL+KCCA and the data after CoKL+KCCA into 2 dimensions (See Fig.~\ref{fig:project}).
The black square, red circle and blue cross represent three different classes. 
Specifically, Fig.~\ref{fig:project}a and Fig.~\ref{fig:project}b are the projected data points from two incomplete seeds datasets (with missing rate 90\%).
It is important to note that since both the datasets are incomplete,
we first use naive filling strategy (using average value for continuous features and majority value for discrete features) 
to complete the datasets in order to get initial kernel matrices. 
Then we apply KCCA on both the initial kernel matrices before CoKL and the complete kernel matrices after CoKL.
The shown dimensions are generated by applying PCA on the results of KCCA.
As compared with Fig.~\ref{fig:project}(a) and Fig.~\ref{fig:project}(b), 
it can be clearly observed in Fig.~\ref{fig:project}(c) that the data is more separable after CoKL.

\begin{figure}
\centering
\includegraphics[width=0.95\columnwidth]{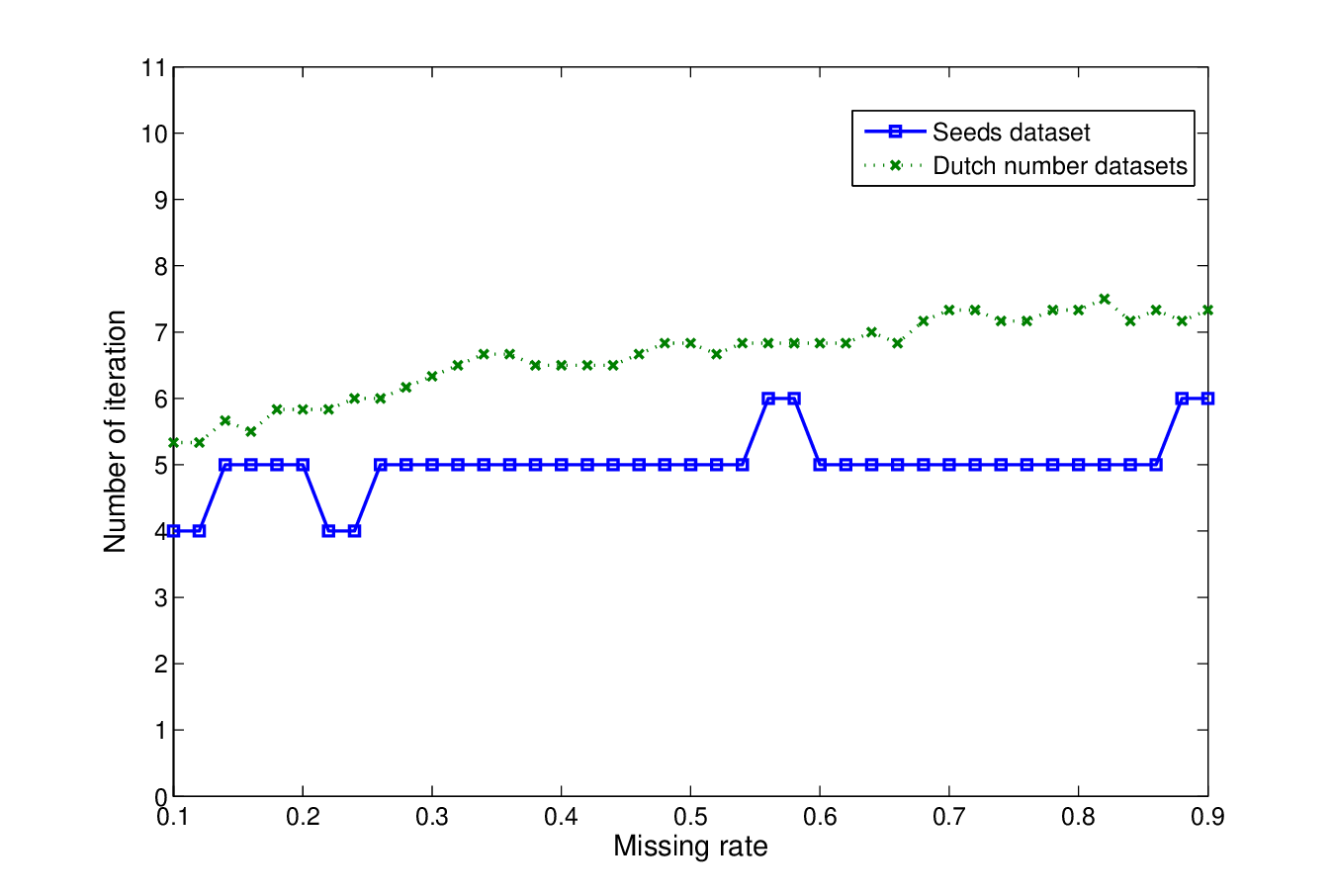}
\caption{The average number of iterations to converge for different missing rates.}
\label{fig:iteration}
\end{figure}
In order to show how the missing rate affects the convergence rate, 
we record the number of iterations to convergence for different missing rates. 
Since we have six pairs of datasets for the handwritten Dutch numbers recognition experiment, 
we take the average number of iterations to converge of the six pairs. 
The result is shown in Fig~\ref{fig:iteration}.
As it can be observed, CoKL takes less than 10 iterations to converge for different experiments and different missing rates.
From the figures, we can also conclude that as the missing rate becomes larger, i.e., more instances missing from the datasets, 
the number of iterations needed to get convergence become slightly larger. 
In fact, the number of iterations needed to get convergence grows really slowly. 
For the smaller dataset, UCI seeds dataset, it is almost constant. 
For the handwritten Dutch numbers recognition experiment, it grows linearly w.r.t. the missing rate with a very small increase rate.

%All the above experiments are done for two incomplete datasets. 
%However, the CoKL also works on more than two incomplete datasets.
%To show the effectiveness of CoKL on more than two incomplete datasets, 
%three different datasets in Handwritten Dutch Numbers Recognition are selected. 
%We randomly delete the instances in all three of the datasets to make them incomplete.
%We run the proposed clustering algorithm for different missing rate (from 10\% to 90\%)  
%Fig~\ref{fig:three} shows the results of clustering. As it can be observed
\section{Related Work}
There are several areas of related works upon which the proposed model is built. First, multi-view learning \cite{Blum:1998:CLU:279943.279962, DBLP:conf/sdm/LongYZ08,
Nigam:2000:AEA:354756.354805,
Kriegel:2008:MMS:1546682.1547113,
Hardoon:2004:CCA:1119696.1119703,
NIPS2011_0817}, is proposed to learn from instances which have multiple representations in different feature space. 
For example, \cite{DBLP:conf/icdm/BickelS04} developed and studied partitioning and agglomerative, hierarchical multi-view clustering algorithms for text data. 
\cite{NIPS2011_0817,ICML2011Kumar_272} are among the first works proposed to solve the multi-view clustering problem via spectral projection.
\cite{DBLP:conf/sdm/LongYZ08} proposed a novel approach to use mapping function to make the clusters from different pattern spaces comparable and hence an optimal cluster can be learned from the multiple patterns of multiple views. 
Linked Matrix Factorization \cite{TangLD09}  is proposed to explore clustering of a set of entities given multiple graphs. It is among the first works proposed to solve the special case that each view is a graph.
Recently, heterogeneous learning \cite{xiaoxiao2012} is proposed to perform clustering where some of the views are graphs and some contain 
vector-based features, however the focus is different from ours.

Another area of related work is learning with incomplete data \cite{Ghahramani94supervisedlearning, Ghahramani95learningfrom, Zhang:2003:CID:964564.964573, Hathaway:2002:CIR:633720.633734, trivedimultiview, Dempster77maximumlikelihood}. 
For example, both of \cite{Zhang:2003:CID:964564.964573} and \cite{Hathaway:2002:CIR:633720.633734} use fuzzy c-means algorithm to cluster incomplete data. 
\cite{Ghahramani94supervisedlearning} is among the first works for learning from incomplete data using EM algorithm for density estimates of mixture models. 
\cite{Hathaway:2002:CIR:633720.633734}  uses a simple triangle inequality-based approximation scheme and applies non-Euclidean relational fuzzy c-means algorithm, while \cite{Zhang:2003:CID:964564.964573} takes advantage of the robustness of kernel fuzzy c-means algorithm. 
However, the focus of these works is different from ours.
Recently, \cite{trivedimultiview} proposed a novel kernel based approach which allows clustering algorithms to be applicable 
when only one (the primary) view is complete, and the other views are incomplete.

There are some differences between our work and the previous approaches.  
First, so far as we know, all of the previous works could not deal with incomplete datasets or at least requires one primary dataset to be complete. 
CoKL works for problems where even no complete datasets are available, which could not be solved by the previous works. 
Second, CoKL completes the kernel matrices collectively and efficiently. CoKL converges really fast, and the number of iterations needed to get convergence does not change too much for different missing rates. 

\section{Conclusion}
In this paper, we study the problem of clustering for multiple incomplete datasets. 
We propose a CoKL principle to deal with the incompleteness of the datasets by  
collectively completing the kernel matrices of the datasets using the common instances in different datasets.
An optimization problem is derived from the CoKL principle to optimize the alignment of incomplete kernel matrices,
and an approximation solution is obtained by iteratively solving a constrained optimization problem.
Furthermore, we propose a clustering algorithm using CoKL and KCCA. 
By applying KCCA after CoKL, the proposed algorithm could maximize the correlation between the projected feature spaces, 
which will increase the performance of clustering compared with other methods.
Two sets of experiments were performed to evaluate the clustering algorithm.
It can be clearly observed that the proposed algorithm outperforms the comparison algorithms 
by as much as twice in NMI. 
We also analyze the efficiency of CoKL, i.e., the number of iterations needed to get convergence. 
The result shows CoKL converges very fast (at most 10 iterations). 
The result on different missing rates shows that the number of iterations needed to get convergence grows linearly with the missing rate at a very small increase rate.

\textbf{Acknowledgement} This work is supported in part by NSF through grants CNS-1115234,  DBI-0960443, and OISE-1129076,  US  Department of Army through grant W911NF-12-1-0066, and Huawei Grant.
%\bibliographystyle{IEEEtran}
\bibliographystyle{IEEEtran}
\bibliography{IEEEabrv,mybib}

%\begin{thebibliography}{1}
%\end{thebibliography}

\end{document}